%% file: DSSN.tex
\documentclass[sigconf,authorversion]{acmart}
\acmSubmissionID{969}

\usepackage{amsfonts}
\usepackage{amsmath}
\usepackage{booktabs} 
\usepackage{cancel}
\usepackage{enumitem}
\usepackage{graphicx}
\usepackage{makecell}
\usepackage{multirow}
\usepackage{subcaption}
\usepackage{textcomp}
\usepackage{xcolor}

\usepackage[ruled]{algorithm2e} 

\SetAlFnt{\small}
\SetAlCapFnt{\small}
\SetAlCapNameFnt{\small}
\SetAlCapHSkip{0pt}

\acmJournal{TOG}

\copyrightyear{2024}
\acmYear{2024}
\setcopyright{rightsretained}
\acmConference[SIGGRAPH Conference Papers '24]{Special Interest Group on Computer Graphics and Interactive Techniques Conference Conference Papers '24}{July 27-August 1, 2024}{Denver, CO, USA}
\acmBooktitle{Special Interest Group on Computer Graphics and Interactive Techniques Conference Conference Papers '24 (SIGGRAPH Conference Papers '24), July 27-August 1, 2024, Denver, CO, USA}
\acmDOI{10.1145/3641519.3657498}
\acmISBN{979-8-4007-0525-0/24/07}

\usepackage[capitalize]{cleveref}
\crefname{section}{Sec.}{Secs.}
\Crefname{section}{Section}{Sections}
\Crefname{table}{Table}{Tables}
\crefname{table}{Tab.}{Tabs.}

\usepackage{scalerel}

\usepackage{pifont}
\newcommand{\cmark}{\textcolor{teal}{\ding{51}}}%
\newcommand{\xmark}{\textcolor{red}{\ding{55}}}%

\newcommand{\flowerbrackets}[1]{\left\{ #1 \right\}}
\newcommand{\commonbrackets}[1]{\left( #1 \right)}

\newcommand{\mlpm}{\mathbf{M}}
\newcommand{\gridg}{\mathbf{G}}
\newcommand{\planes}{\mathbf{S}}

\newcommand{\tensorfscene}{\mathcal{F}_{s}}
\newcommand{\tensorfflow}{\mathcal{F}_{f}}

\newcommand{\pixelq}{\mathbf{q}}
\newcommand{\pointp}{\mathbf{p}}
\newcommand{\viewv}{\mathbf{v}}

\newcommand{\colorc}{\mathbf{c}}
\newcommand{\opticalflowf}{\mathbf{f}}

\newcommand{\featureh}{\mathbf{h}}

\newcommand{\sparseFlowPrior}{\mathcal{P}_{\text{sf}}}
\newcommand{\denseFlowPrior}{\mathcal{P}_{\text{df}}}
\newcommand{\sparseDepthPrior}{\mathcal{P}_{\text{sd}}}
\newcommand{\loss}{{\mathcal{L}}}
\newcommand{\losscolor}{{\mathcal{L}_{\text{ph}}}}
\newcommand{\losssd}{{\mathcal{L}_{\text{sd}}}}

\newcommand{\losssf}{{\mathcal{L}_{\text{sf}}}}
\newcommand{\lossdf}{{\mathcal{L}_{\text{df}}}}

\begin{document}
    \title{Factorized Motion Fields for Fast Sparse Input Dynamic View Synthesis}

\author{Nagabhushan Somraj}
\orcid{0000-0002-2266-759X}
\affiliation{%
  \institution{Indian Institute of Science}
  \city{Bengaluru}
  \state{Karnataka}
  \postcode{560012}
  \country{India}}
\email{nagabhushans@iisc.ac.in}
\author{Kapil Choudhary}
\orcid{0009-0000-2643-6329}
\affiliation{%
  \institution{Indian Institute of Science}
  \city{Bengaluru}
  \country{India}
}
\email{kapilc@iisc.ac.in}
\author{Sai Harsha Mupparaju}
\orcid{0009-0003-1638-7051}
\affiliation{%
  \institution{Indian Institute of Science}
  \city{Bengaluru}
  \country{India}
}
\email{saiharsham@iisc.ac.in}
\author{Rajiv Soundararajan}
\orcid{0000-0001-5767-5373}
\affiliation{%
  \institution{Indian Institute of Science}
  \city{Bengaluru}
  \country{India}}
\email{rajivs@iisc.ac.in}

    \begin{abstract}
        Designing a 3D representation of a dynamic scene for fast optimization and rendering is a challenging task.
        While recent explicit representations enable fast learning and rendering of dynamic radiance fields, they require a dense set of input viewpoints.
        In this work, we focus on learning a fast representation for dynamic radiance fields with sparse input viewpoints.
        However, the optimization with sparse input is under-constrained and necessitates the use of motion priors to constrain the learning.
        Existing fast dynamic scene models do not explicitly model the motion, making them difficult to be constrained with motion priors.
        We design an explicit motion model as a factorized 4D representation that is fast and can exploit the spatio-temporal correlation of the motion field.
        We then introduce reliable flow priors including a combination of sparse flow priors across cameras and dense flow priors within cameras to regularize our motion model.
        Our model is fast, compact and achieves very good performance on popular multi-view dynamic scene datasets with sparse input viewpoints.
        The source code for our model can be found on our project page: \url{https://nagabhushansn95.github.io/publications/2024/RF-DeRF.html}.
    \end{abstract}

%
%

\ccsdesc[500]{Computing methodologies~Rendering}
\ccsdesc[500]{Computing methodologies~Volumetric models}
\ccsdesc[300]{Computing methodologies~Computer vision}
\ccsdesc[300]{Computing methodologies~Computational photography}
\ccsdesc[300]{Computing methodologies~Regularization}

%
%

    \keywords{Fast dynamic view synthesis, dynamic radiance fields, factorized models, motion priors, sparse input views}

    \maketitle

    \section{Introduction}\label{sec:introduction}
    Synthesizing novel views of a scene given images/videos of the scene in other viewpoints has several applications such as meta-verse, sports, and remote exploration.
    Neural Radiance Fields (NeRF) \cite{mildenhall2020nerf} brought in a seminal shift in novel view synthesis by incorporating differential volume rendering with a compact continuous-depth model.
    However, it has several limitations such as the inability to handle object motion in the scene as well as long training and rendering times.
    Recently, K-Planes~\cite{fridovich2023kplanes} significantly reduced the time complexity for dynamic view synthesis by proposing a factorized 4D volume representation.
    However, K-Planes requires a large number of input viewpoints to render photo-realistic novel views when employing a multi-view camera setup.
    The goal of our work is to overcome this limitation and design a fast dynamic radiance field that can effectively learn the dynamic scene with few input viewpoints.
    We mainly focus on a sparse multi view camera setting, where a video from each viewpoint is available.

    Optimizing a dynamic radiance field with few input viewpoints is highly under-constrained.
    A popular approach to regularize an under-constrained system is to impose additional priors during the optimization.
    For example, sparse static radiance fields are regularized with depth~\cite{deng2022dsnerf,uy2023scade}, visibility~\cite{somraj2023vipnerf} and natural image~\cite{wu2024reconfusion} priors.
    However, the object motion in dynamic scenes motivates the study of motion priors for sparse input dynamic view synthesis.
    Since K-Planes employs a 4D volumetric representation without a motion model, it does not allow the motion implicitly learned by the model to be regularized using motion priors.
    Thus, there is a need to design a dynamic radiance field with an explicit motion field that lends itself to be constrained with motion supervision.

    We design a dynamic radiance field consisting of two models, a 5D radiance field that learns the 3D scene at a canonical time instant and a 4D motion or deformation field that learns the motion from any time instant $t$ to the canonical time instant $t'$.
    Since the motion model is a one-directional mapping from $t$ to $t'$, it is not obvious how to impose flow priors across two arbitrary time instants.
    We achieve this by constraining our motion model to map a pair of matched points, obtained using the motion prior, to the same 3D point in the canonical volume.
    \cref{fig:loss-flow} illustrates our approach of imposing the motion prior.
    This allows us to impose motion priors across any two time instants and across any cameras.

    Prior and concurrent works~\cite{fang2022tineuvox,shaw2023swags} employ deep neural networks (DNN) to learn the motion field.
    However, the use of the DNN to learn the motion makes the model computationally expensive.
    The challenge here is to design a motion model that can learn the motion efficiently while yielding fast training and rendering.
    While explicit models such as voxel grids~\cite{fridovich2022plenoxels,sun2022dvgo}, hash-grids~\cite{muller2022instant} and 3D Gaussians (3DGS)~\cite{kerbl20233dgs} are shown to be effective in learning static scenes, naively extending these techniques to model 4D motion may not be efficient.
    For example, extending 3D voxel grids to 4D scales the memory requirement to the fourth power of grid resolution.
    Further, for a given object, since the motion exists at every time instant, models that exploit scene sparsity to reduce the memory requirement may not extend effectively to 4D\@.
    Since the motion is dense and has a high spatio-temporal correlation, we employ factorized volumes to exploit the correlation.
    Specifically, we employ a 4D factorized volume to learn the scene flow from any time $t$ to the canonical time $t'$.

    An ideal motion prior to regularize the motion field is a dense flow prior across cameras and time instants.
    Prior models such as NSFF~\cite{li2021nsff} use dense optical flow to supervise an auxiliary task of predicting the flow between two frames, which may not efficiently regularize the core dynamic radiance field model.
    Further, we observe that deep flow estimation networks used in such works are sensitive to variations in camera parameters, lighting changes and occlusions, and hence find it challenging to determine reliable dense correspondences across cameras.
    Regularizing the model with such noisy flow priors may lead to sub-optimal performance.

    We address these challenges by employing a combination of two complementary flow priors, a rich but sparse flow prior across cameras and a dense flow prior within individual cameras.
    In particular, we obtain reliable flow priors across cameras and time instants for a sparse set of SIFT keypoints.
    Within individual cameras, we obtain reliable dense flow priors using deep optical flow estimation networks~\cite{teed2020raft}.
    These priors within the same camera tend to be more reliable owing to smaller variations in camera parameters and lighting changes.
    The reliability of our priors as opposed to the dense flow priors across cameras is illustrated in \cref{fig:failure-dense-flow}.
    The combination of the sparse across camera flow priors and the dense within camera flow priors provides an effective alternative to the cross camera dense flow priors.

    We evaluate our model on two popular multi-view dynamic scene datasets and find that our model outperforms the state-of-the-art dynamic view synthesis models with fewer input viewpoints.
    We refer to our model as RF-DeRF, since we employ Reliable Flow priors for Deformable Radiance Fields.
    We summarize the main contributions of our work in the following:
    \begin{itemize}
        \item We design a fast and compact dynamic radiance field for sparse input dynamic view synthesis by employing an explicit motion model that can be easily regularized using motion priors.
        We employ a 4D factorized volume to exploit the spatio-temporal correlation of the motion field.
        \item We propose a complimentary set of reliable flow priors to regularize the motion field when training with few multi-view videos.
        We obtain a sparse flow prior that regularizes flow across cameras and a dense flow prior that regularizes the flow within individual cameras.
        \item We achieve very good dynamic view synthesis performance on two popular multi-view datasets with very few views.
    \end{itemize}

    \input{tex/tables/related_work}

    \section{Related Work}\label{sec:related-work}
    \subsection{Classical Work on Deformation Models}\label{subsec:classical-work-deformation-models}
    The modeling of dynamic radiance fields through a static radiance field and a motion field is similar to the use of a deformation model that deforms a canonical representation of an entity.
    This is a popular approach to model deformable solids~\cite{sederberg1986free}, human motion~\cite{loper2015smpl}, facial expressions~\cite{bradley2010high}, deformable garments~\cite{miguel2012data,pons2017clothcap}, fluids~\cite{adrian1991particle,gregson2012stochastic,hawkins2005acquisition}, gases~\cite{atcheson2008time}, smoke~\cite{hawkins2005acquisition}, flames~\cite{ihrke2004image} as well as CT (Computed Tomography) and MRI (Magnetic Resonance Imaging) scans in the medical field~\cite{nunes2012imaging}.
    Learning the motion or deformation field is also found to improve the reconstruction of 3D scenes~\cite{vedula2000shape} and free-viewpoint rendering of dynamic scenes~\cite{carranza2003free}.

    The canonical space in such deformable models is popularly represented as triangular meshes estimated through a multi-view stereo algorithm~\cite{bradley2008markerless,bradley2010high,kraevoy2005template,allen2003space}, or specialized parametric meshes for humans and faces~\cite{anguelov2005scape,vlasic2008articulated,loper2015smpl}, or sums of Gaussians~\cite{stoll2011fast}.
    In contrast, we employ a radiance field to learn the canonical space owing to its advantages over meshes~\cite{mildenhall2020nerf}.
    Different deformation models include free-form deformation (FFD) fields or optical flows~\cite{faloutsos1997dynamic,wang2009physically}, bijective mappings~\cite{kraevoy2004cross}, morphable models~\cite{blanz1999morphable}, splines~\cite{vlasic2009dynamic}, volumetric Laplacian deformations~\cite{de2008performance} and linear systems with basis functions~\cite{ihrke2004image}.
    Our motion field is closer to the free-form deformation among the above deformation approaches, but differs in the representation used for the motion field.
    Finally, the sparse flow priors we employ can be thought of as similar to tracking the markers~\cite{guskov2003trackable,hasler2006physically} or keypoints~\cite{bradley2008markerless,joo2018total} while estimating the classical deformation models.

    \subsection{Dynamic View Synthesis}\label{subsec:dynamic-view-synthesis}
    Different from the use of deformable models, \citet{zitnick2004high} learn the 3D scene dynamics using a layered depth representation with motion compensation.
    More recent volumetric representations such as Multiplane Images (MPI)~\cite{zhou2018stereomag} are extended to handle dynamic scenes by employing an MPI per frame~\cite{lin2021deep} or temporal basis functions~\cite{xing2021temporal}.
    However, such approaches suffer from depth discretization artifacts~\cite{mildenhall2020nerf}.
    Instead of volumetric models, \citet{yoon2020novel} employ depth image based warping, where the depth is obtained by combining monodepth with multi-view depth.
    Recent work on dynamic view synthesis with sparse input viewpoints require depth information obtained through the use of RGB-D cameras~\cite{li2023dynamic} or multi-view stereo~\cite{bansal2023npc}.
    Further, such approaches struggle to handle soft edges and translucent objects~\cite{penner2017soft3d}.

    \subsection{Dynamic Radiance Fields}\label{subsec:dynamic-radiance-fields}
    In contrast to the classical approaches, dynamic view synthesis can be solved by learning a 6D radiance field that maps the position, time and viewing direction to the radiance.
    However, the performance of such approaches degrades when the input viewpoints are sparse.
    We provide a quick overview of prior work that share a few attributes of our model in \cref{tab:related-work}.
    To the best of our knowledge, ours is the first work to address dynamic radiance fields with few input viewpoints while achieving fast training and rendering.
    In the following, we review the literature on dynamic radiance fields and sparse-input static radiance fields.

    Dynamic radiance fields can be broadly classified into two categories based on how the temporal modeling is handled, which is crucial in the sparse input setting.
    A simple approach is to model the dynamic radiance field as a 6D function of position, time, and viewing direction.
    To utilize flow priors, NSFF~\cite{li2021nsff} predicts flow as an auxiliary task and constrains the volume density of the mapped 3D points provided by the flow.
    K-Planes~\cite{fridovich2023kplanes} and HexPlane~\cite{cao2023hexplane} extend TensoRF~\cite{chen2022tensorf} to a 4D model that maps the position and time to a latent feature, which is then decoded by a tiny multi-layer perceptron (MLP).
    Instead of conditioning the radiance field directly on time, DyNeRF~\cite{li2022neural} conditions the radiance field on a per-time-instant feature vector which is jointly optimized with the radiance field.
    The lack of a motion model in these approaches makes it incompatible to impose motion priors when learning with sparse input viewpoints.

    The second set of models employs a motion or deformation field that maps the 3D points from a given time instant to a canonical time instant~\cite{pumarola2021dnerf,wang2023flow,wang2023tracking}.
    TiNeuVox~\cite{fang2022tineuvox}, SWAGS~\cite{shaw2023swags} and CoGS~\cite{yu2023cogs} replace the scene representation MLP in D-NeRF~\cite{pumarola2021dnerf} with a TensoRF or 3DGS model, but use MLPs to model the motion field, leading to expensive training and rendering time.
    To achieve fast optimization and rendering, DeVRF~\cite{liu2022devrf} employs a 4D voxel grid to learn the deformation field, but its memory requirements scale with the fourth power of the grid resolution.
    Instead of employing a free-form deformation model, \citet{wang2021neural} models the motion using discrete cosine transform (DCT) basis functions.
    Different from the above, prior~\cite{guo2023forward} and concurrent work~\cite{wu20234dgs} employ a motion model that maps the 3D points from the canonical time instant to a given time instant.
    However, the forward warping of points can lead to holes in novel views, which may need to be infilled using a separate inpainting network~\cite{guo2023forward,mirzaei2023spinnerf,somraj2022decompnet}.

    \subsection{Radiance Fields with Sparse Views}\label{subsec:radiance-fields-with-sparse-views}
    Optimizing a radiance field with sparse input views for static scenes has rich literature.
    With sparse inputs, the volume rendering equations are severely under-constrained leading to distortions such as floaters.
    Few-shot NeRF models address this by constraining the model further with hand-crafted or learned priors.
    While the early works employ simple priors such as sparsity of mass~\cite{kim2022infonerf} and smoothness of depth~\cite{niemeyer2022regnerf}, later works found that the use of richer priors can lead to better performance.
    The most popular prior is the depth prior~\cite{deng2022dsnerf,roessle2022ddpnerf,uy2023scade,somraj2023simplenerf}, while other priors based on visibility~\cite{somraj2023vipnerf}, semantic information~\cite{jain2021dietnerf} and diffusion models~\cite{wynn2023diffusionerf,wu2024reconfusion} are also explored.
    While most of the works employ the NeRF baseline leading to long training time, recent works explore the use of explicit models~\cite{shi2024zerorf,xiong2023sparsegs,zhu2023fsgs}.
    However, none of these approaches analyze or regularize the distortions caused by motion in a dynamic scene.

    \input{tex/figures/model_architecture}

    \subsection{Fast Optimization of Radiance Fields}\label{subsec:fast-optimization-of-radiance-fields}
    Fast optimization of radiance fields allows wider usage in real-world applications.
    Voxel grid based approaches trade memory for speed by replacing the MLP in NeRF with a voxel grid~\cite{fridovich2022plenoxels,sun2022dvgo}.
    Prior work on fast and compact representation for learning radiance fields can be broadly classified into two categories.
    One category of models employs a factorized volume representation~\cite{chen2022tensorf,fridovich2023kplanes,cao2023hexplane} to exploit the spatial correlation of the entities to be learned, such as volume density and color.
    The other category of models~\cite{kerbl20233dgs,muller2022instant,barron2023zipnerf} mainly exploit the sparsity of the scene to reduce the memory footprint.
    However, the scene flow for a moving object is non-zero at every time instant, but its variation is temporally correlated.
    Since the motion field is not sparse temporally, we believe that the sparsification based approaches may tend to learn an independent motion field for every time instant.
    For example, with a 4DGS model, a single 4D Gaussian may not be able to model the motion across a few time instants and can model the motion at a single time instant only, similar to \citet{lee2023fast}.
    On the other hand, factorized volumes can effectively exploit the spatio-temporal correlation of the motion field and hence we employ a 4D factorized model to learn the motion field.

    \section{Method}\label{sec:method}
    Given a set of posed video frames $I_t^v$, where $t=\left\{ 1, 2, \ldots, N_f \right\}$ denotes the time instant or the frame index among $N_f$ frames per camera and $v=\left\{ 1, 2, \ldots N_c \right\}$ denotes the camera index among $N_c$ stationary cameras, the goal is to synthesize the dynamic scene at a specified time instant $t=\flowerbrackets{1, 2, \ldots, N_f}$ and in any novel view.
    We focus on the sparse multi-view setting, where the number of cameras $N_c$ is small.
    The main challenges here include the design of an explicit motion model that allows regularization with motion priors and the choice of reliable motion priors itself when training with few input views.
    We first describe the design of our dynamic radiance field (\cref{subsec:base-derf}), and then discuss its training with reliable flow priors (\cref{subsec:rf-derf}).

    \subsection{Dynamic Radiance Field}\label{subsec:base-derf}
    We modify the K-Planes~\cite{fridovich2023kplanes} implementation for dynamic radiance fields to incorporate an explicit motion model.
    K-Planes employs a multi-resolution grid $\gridg$ followed by a tiny multi-layer perceptron (MLP) $\mlpm$ to represent a radiance field.
    As shown in \cref{fig:model-architecture}, we model our dynamic radiance field by employing two factorized tensorial models, a radiance field $\tensorfscene = \gridg_s \circ \mlpm_s$ that represents the 3D scene at a canonical time instant $t'$, and a 4D scene flow or deformation field $\tensorfflow = \gridg_f \circ \mlpm_f$ that represents the scene flow for a 3D point $\pointp$ from time $t$ to $t'$.
    Mapping the 3D scene at every time instant to a canonical volume helps our model enforce temporal consistency of objects in the scene~\cite{pumarola2021dnerf}.

    Here, $\gridg_f$ consists of six planes $\{ \planes_{xy}, \planes_{yz}, \planes_{xz}, \planes_{xt}, \planes_{yt}, \planes_{zt} \}$ at every resolution, where the first three planes model the spatial correlation and the next three model the spatio-temporal correlation of the motion field.
    To obtain the scene flow for a 3D point $\pointp$ from time $t$ to $t'$, we first project the 4D point $(\pointp, t)$ onto the six planes and bilinearly interpolate the feature vectors in each of the six planes.
    We then combine the features using the Hadamard product to obtain the final feature vector $\featureh_i$ as
    \begin{align}
        \featureh_i = \gridg_f \commonbrackets{\pointp} &= \prod_{c \in \{ xy, yz, xz, xt, yt, zt \}} S_c \commonbrackets{\pointp},
        \label{eq:factorized-grid-flow}
    \end{align}
    which is then fed to the tiny MLP $\mlpm_f$ that outputs the scene flow.
    Modeling the scene flow field $\tensorfflow$ using the hex-plane representation and the tiny MLP makes our motion model fast and compact.
    For more details on the hex-plane representation, we refer the readers to K-Planes~\cite{fridovich2023kplanes}.
    We model the scene grid $\gridg_s$ using a similar factorization with three spatial planes.
    While it is beneficial to model the motion field using a factorized volume, the canonical scene can be learned using any explicit representation such as 3DGS~\cite{kerbl20233dgs}.
    We employ factorized volumes to model both scene and motion using a unified framework.

    We train our model, similar to K-Planes, by rendering randomly sampled pixels $\pixelq_t^v$ in view $v$ at time $t$ and using the ground truth color as supervision.
    To render a pixel $\pixelq_t^v$, we sample $N_p$ points $\flowerbrackets{\pointp_i}_{i=1}^{N_p}$ at depths $\flowerbrackets{z_i}_{i=1}^{N_p}$ along the corresponding ray.
    For every 3D point $\pointp_i$, we first obtain the corresponding 3D point $\pointp_i'$ at canonical time $t'$ by computing the scene flow from time $t$ to $t'$ using $\tensorfflow$ as
    \begin{align}
        \pointp_i' &= \tensorfflow \commonbrackets{\pointp_i, t} + \pointp_i. \label{eq:flow-mapping}
    \end{align}
    We then query $\gridg_s$ at $\pointp_i'$ to obtain the volume density $\sigma_i$ and a latent feature $\featureh_i'$ corresponding to $\pointp_i$ as
    \begin{align}
        \sigma_i, \featureh_i' = \gridg_s \commonbrackets{\pointp_i'}.
    \end{align}
    A tiny MLP $\mlpm_s$ maps $\featureh_i'$, encoded viewing direction $\viewv$ and encoded time $t$ to the color $\colorc_i$ of $\pointp_i$ as
    \begin{align}
        \colorc_i = \mlpm_s \commonbrackets{\featureh_i', \gamma\commonbrackets{\viewv}, \gamma\commonbrackets{t}},
    \end{align}
    where $\gamma$ denotes the encoding~\cite{muller2022instant} of the viewing direction and the time instant.
    Conditioning $\mlpm_s$ additionally on time allows our model to capture illumination changes due to object motion.
    $\colorc_i$ are then volume rendered to obtain the color $\colorc$ of $\pixelq$ as
    \begin{align}
        w_i &= \commonbrackets{\Pi_{j=1}^{i-1}\exp\commonbrackets{-\delta_j \sigma_j}} \commonbrackets{1 - \exp\commonbrackets{-\delta_i \sigma_i}}, \label{eq:volume-rendering1} \\
        \colorc &= \sum_{i=1}^{N_p} w_i \colorc_i, \quad z = \sum_{i=1}^{N_p} w_i z_i, \label{eq:volume-rendering2}
    \end{align}
    where $\delta_i = z_i - z_{i-1}$ and $z$ gives the expected depth of $\pixelq_t^v$.
    Both $\tensorfflow$ and $\tensorfscene$ are optimized through the photometric loss, $\losscolor = \Vert \colorc - \hat{\colorc} \Vert^2$, where $\hat{\colorc}$ is the ground truth color.
    Decomposing the scene into a canonical scene field and a flow field allows us to regularize the flow field using flow priors when only a sparse set of viewpoints are available.

    \input{tex/figures/loss_flow}

    \subsection{Training with Flow Priors}\label{subsec:rf-derf}

    \input{tex/figures/flow_matches}

    \input{tex/tables/quantitative_main}

    A dense flow prior across cameras and across time instants is an ideal prior for optimal regularization of our motion field.
    However, different cameras in a multi-view setting could suffer from variations in parameters such as intrinsics, color balance, exposure and focus making it challenging to determine correspondences across cameras based on luminance constancy.
    As a result, the dense flow estimates obtained using deep optical flow networks are unreliable across cameras as seen in \cref{fig:cross-view-dense-flow}.
    On the other hand, matching keypoints using robust SIFT~\cite{lowe2004sift} descriptors are more reliable across cameras as seen in \cref{fig:cross-view-sparse-flow}.
    Specifically, for a pair of frames across time instants $t$ and $s$ and cameras $v$ and $u$, we extract the SIFT keypoints in individual images using Colmap~\cite{schonberger2016colmap} and match the keypoints using SIFT descriptors.
    The difference between the locations of the matched keypoints gives the sparse flow prior $\sparseFlowPrior$.
    While Colmap provides a reasonable number of matches, newer models such as R2D2~\cite{revaud2019r2d2} could be employed to improve the richness of the sparse flow prior.

    Employing flow priors from any time $t$ to canonical time $t'$ for large $t - t'$ may be less efficient since large regions of the scene may not be visible.
    Thus, it is desirable to utilize flow priors across a short duration of time.
    However, the motion model $\tensorfflow$ provides the scene flow from $t$ to $t'$ only and not the flow between any two arbitrary time instants.
    We resolve this by using the flow prior to encourage $\tensorfflow$ to map a pair of matched points at different time instants to the same object location in 3D\@.
    Specifically, we obtain the matching pixels using the flow priors and constrain the 3D points corresponding to the matching pixels to map to the same 3D point in the canonical scene field as shown in \cref{fig:loss-flow}.

    Mathematically, let the flow corresponding to a pixel $\pixelq_t^v$ to time $s$ and view $u$ be given by $\opticalflowf_{t \rightarrow s}^{v \rightarrow u}$.
    Then the matching pixel at time $s$ and view $u$ is given by $\pixelq_s^u = \pixelq_t^v + \opticalflowf_{t \rightarrow s}^{v \rightarrow u}$.
    Let the 3D points sampled along the rays corresponding to $\pixelq_t^v$ and $\pixelq_s^u$ be given by $\pointp_i(t,v)$ and $\pointp_i(s,u)$ respectively.
    Then, we impose the sparse flow constraint as
    \begin{align}
        \losssf &= \left\Vert \sum_{i=0}^{N_p-1} w_i(t,v) \pointp_i'(t,v) - \sum_{i=0}^{N_p-1} w_i(s,u) \pointp_i'(s,u) \right\Vert^2, \label{eq:loss-flow}
    \end{align}
    where $\pointp'(t,v)$ and $\pointp'(s,u)$ are computed from $\pointp(t,v)$ and $\pointp(s,u)$ using $\tensorfflow$ as in \cref{eq:flow-mapping}, and $w_i(t,v)$ and $w_i(s,u)$ are computed using \cref{eq:volume-rendering1}.
    The two terms in \cref{eq:loss-flow} represent $\pixelq_t^v$ and $\pixelq_s^u$ in the canonical volume respectively.
    Thus, $\losssf$ guides $\tensorfscene$ on which two points in two different time instants belong to the same object.

    In \cref{eq:loss-flow}, we first find the canonical field points $\pointp_i'$ using $\tensorfflow$ and then average the location of the 3D points using weights $w_i$ and not the other way for the following reason.
    The former approach regularizes the motion model $\tensorfflow$ for every $\pointp_i$ giving a rich supervision to $\tensorfscene$, whereas the latter approach regularizes $\tensorfflow$ only for the expected 3D point $\pointp=\sum w_i \pointp_i$.
    Further, since we do not impose stop-gradient on $w_i$ in $\losssf$, the flow priors also help remove incorrect masses such as floaters.

    While the sparse flow prior $\sparseFlowPrior$ is rich on account of matching pixels across cameras and time instants, the prior is available at only sparse keypoints.
    Thus, we complement $\sparseFlowPrior$ with a dense flow prior $\denseFlowPrior$ within individual cameras.
    For a pair of frames from the same camera at two closer time instants, there is little variation in the camera intrinsics and pipeline parameters.
    Hence, the dense flow priors obtained using a deep optical flow network are more reliable as seen in \cref{fig:same-view-dense-flow}.
    We employ the popular optical flow estimation model RAFT~\cite{teed2020raft} to obtain the dense flow priors and constrain our model with a dense flow loss $\lossdf$ similar to \cref{eq:loss-flow}.
    We find that the dense flow priors typically help reduce artifacts in static regions.

    We train our model by minimizing the combination of photometric loss $\losscolor$, sparse flow loss $\losssf$ and dense flow loss $\lossdf$ as
    \begin{align}
        \loss &= \losscolor + \lambda_{\text{sf}} \losssf + \lambda_{\text{df}} \lossdf , \label{eq:overall-loss}
    \end{align}
    where $\lambda_{\text{sf}}$ and $\lambda_{\text{df}}$ are hyper-parameters.

    \input{tex/figures/qualitative_main_n3dv_2views}

    \section{Experiments}\label{sec:experiments}
    \subsection{Evaluation Setup}\label{subsec:evaluation-setup}
    \emph{Datasets:}
    We evaluate our model on two popular multi-view dynamic scene datasets, namely N3DV~\cite{li2022neural} and InterDigital~\cite{sabater2017interdigital} with two, three, and four input views.
    Following prior work~\cite{fridovich2023kplanes}, we downsample the videos spatially by a factor of two for all the experiments.
    We use the video at the center of the camera rig for testing and uniformly sample train videos from the remaining videos.
    \emph{N3DV dataset} contains six real-world scenes with 17--21 static cameras per scene and 300 frames per viewpoint.
    The videos have a spatial resolution of $1352 \times 1014$ and a frame rate of 30fps.
    \emph{InterDigital dataset} contains multiple real-world scenes with 16 static cameras and varying number of frames per scene.
    We undistort the video frames using the radial distortion parameters provided with the dataset and use the undistorted videos for our experiments.
    We select five scenes that contain at least 300 frames and choose the first 300 frames.
    The videos have a spatial resolution of $1024 \times 544$ and a frame rate of 30fps spanning 10 seconds in all our experiments.

    \emph{Evaluation measures:}
    We evaluate the rendered frames of all the methods using PSNR, SSIM~\cite{wang2003multiscale} and LPIPS~\cite{zhang2018unreasonable}.
    We also evaluate the models on their ability to reconstruct the 3D scene by computing MAE on the rendered depth maps.
    Due to the unavailability of true depth maps on both datasets, we use the depth provided by K-Planes trained with dense input views as pseudo ground truth depth.

\input{tex/tables/quantitative_ablations}

    \subsection{Comparisons and Implementation Details}\label{subsec:comparisons}
    We mainly compare the performance of our model against K-Planes and HexPlane.
    We use the official code released by the authors of K-Planes and modify it to implement our model.
    We refer to our base model without any priors as DeRF model.
    To test the superiority of our priors, we also compare our model against K-Planes with sparse depth priors and DeRF with dense flow priors across cameras.
    Since we use two factorized models instead of one used in K-Planes, we reduce the feature dimension in both of our models by half to keep the total number of parameters comparable to K-Planes.
    We train all the models for 30k iterations on a single NVIDIA RTX 2080 GPU\@.
    We randomly pick $s \in \{t-10, t+10\}$ to impose the flow prior losses and set $\lambda_{\text{sf}} = 1$ and $\lambda_{\text{df}} = 1$.
    We use the same values as suggested by K-Planes for all the remaining hyperparameters.

    \input{tex/tables/quantitative_cross_camera_dense_flow}

    \subsection{Results}\label{subsec:results}
    We show the quantitative performance of our model, K-Planes and HexPlane in \cref{tab:quantitative-results-main}, where we observe that our RF-DeRF model outperforms both K-Planes and HexPlane across all the settings on both the datasets.
    We observe that the performance of all the models is relatively higher on the N3DV dataset as compared to the InterDigital dataset.
    This is perhaps due to the InterDigital dataset having larger motion and highly textured regions.
    We also note that the performance of HexPlane is substantially lower than both K-planes and our model.
    This could be a result of optimizing HexPlane initially with a low-resolution grid, which causes the model to overfit the input-views and not utilize the high-resolution grid in the later stages of training.
    K-Planes does not suffer from this drawback by using a multi-resolution grid throughout the optimization.
    From \cref{fig:qualitative-main-n3dv-3views,fig:qualitative-main-n3dv-4views,fig:qualitative-main-id-2views}, we observe that our model is able to correct errors both in moving objects and in static regions over K-Planes.
    In particular, we observe that K-Planes suffer from blur, deformation, and disappearance of moving objects while rendering novel views, which are mitigated by our model.
    The improvements are more starkly visible in the supplementary videos.
    We compare the rendered depth from both the models in \cref{fig:qualitative-depth-id-2views,fig:qualitative-depth-n3dv-3views} and observe that our model is able to reconstruct the depth more accurately than K-Planes.
    \cref{tab:quantitative-results-main} also shows that our DeRF model (without any priors) performs better than K-Planes in the scenes with larger motion, such as the scenes in the InterDigital dataset, perhaps due to the temporal consistency enforced by the canonical volume.
    This is also visible in the example shown in \cref{fig:qualitative-main-n3dv-2views}.
    In our experiments, RF-DeRF roughly took 1.5 hours and 5GB GPU memory to train on a single scene, and 6 seconds to render a single frame.
    The size of the saved model parameters is approximately 280MB\@.
    Please refer to \citet{fridovich2023kplanes} for the training time for various models including K-Planes and DyNeRF\@.

    \input{tex/figures/qualitative_sparse_depth_n3dv_3views}

    \emph{Ablations:}
    We analyze the significance of each component of our model by disabling one component at a time in \cref{tab:quantitative-results-ablations}.
    We observe that the sparse flow prior is the most significant component of our model, giving a large boost in performance when included and a large drop in performance when excluded.
    The effect of sparse flow prior is more pronounced in terms of depth MAE, which measures the accuracy of the reconstructed 3D scene.
    This demonstrates the importance of reliable motion priors across cameras, even if sparse.
    \cref{fig:qualitative-ablations-n3dv-3views} shows that sparse flow prior is more helpful in mitigating motion related artifacts such as motion blur, while within-camera dense flow prior mainly helps reduce distortions in the static regions.
    The combination of the two priors leads to the best performance by mitigating both artifacts.

    \input{tex/tables/quantitative_sparse_depth}

    \emph{Dense cross-camera flow priors vs our priors:}
    To validate our hypothesis that a combination of cross-camera sparse flow and within-camera dense flow priors is more effective than dense flow priors across cameras, we impose cross-camera dense flow priors on our base DeRF model and evaluate the performance on both datasets.
    From \cref{tab:cross-camera-dense-flow} and \cref{fig:qualitative-naive-flow-n3dv-3views}, we observe that imposing noisy cross-camera dense flow priors leads to a large drop in performance, whereas our priors help improve the performance significantly.

    \emph{Depth vs flow priors:}
    We now analyze the effect of flow priors against that of depth priors in dynamic view synthesis with sparse input views.
    The depth priors guide the model on the exact locations of the objects in the scene, but do not account for the correlation of object locations across time instants.
    On the other hand, the flow priors account for such inter-frame correlations as well as encompass the information provided by the depth priors in static regions.
    To experimentally validate our hypothesis, we compare the performance of K-Planes and our DeRF models when regularized with sparse depth priors $\sparseDepthPrior$ obtained through Colmap sparse reconstruction~\cite{deng2022dsnerf}.
    We obtain $\sparseDepthPrior$ for every frame and use them to constrain the K-Planes model as $\losssd = \Vert z - \hat{z} \Vert^2$, where $z$ is the depth obtained in \cref{eq:volume-rendering2} and $\hat{z}$ is the depth prior.
    From \cref{tab:sparse-depth-vs-sparse-flow}, we observe that sparse flow priors lead to a significant improvement in performance as compared to that of sparse depth priors.
    The difference in performance can be more clearly observed in \cref{fig:qualitative-sparse-depth-n3dv-3views} and the supplementary videos.
    This shows the utility of motion priors in dynamic view synthesis with sparse input views.

    \emph{Performance with increasing viewpoints:}
    To understand the impact of our priors as the number of input views increases, we compare the performance of RF-DeRF against K-Planes with different number of cameras on a single scene from N3DV dataset in \cref{fig:quantitative-increasing-views-n3dv}.
    Our model significantly improves the performance over K-Planes in sparse viewpoint scenarios, while matching the performance of K-Planes with dense input views.
    Thus, we conclude that our priors are particularly helpful in improving the reconstruction of under-observed regions in the scene.

    \input{tex/tables/quantitative_dense_views}

    \emph{Performance with dense input views:}
    We now test the ability of our model in reconstructing the dynamic scene when provided with dense multi-view inputs.
    Since motion priors may not be needed when dense views are available, we analyze the performance of our base DeRF model (without any priors).
    From \cref{tab:dense-input-views}, we observe that our DeRF model is competitive with K-Planes.
    Thus, one could employ our model in both dense and sparse input scenarios by additionally employing reliable flow priors in the latter case.

    \subsection{Limitations and Future Work}\label{subsec:limitations}
    Our approach of mapping the scene at every time instant to the canonical volume implies that only the objects present in the canonical volume can be rendered.
    In other words, our model may not handle objects entering or leaving the scene, which could happen over longer durations.
    This could be resolved by learning multiple models, each trained on self-contained shorter duration videos.
    We employ Colmap to generate sparse flow priors in our framework, which is a time-consuming process.
    In our experiments, we found that the generation of sparse flow priors takes about 45 minutes per scene.
    There is a need to explore faster approaches to determine reliable sparse correspondences across cameras.
    Further, our approach requires the cameras to be calibrated and time-synchronized apriori.
    It would be interesting to optimize the scene, camera parameters, and time-synchronization jointly in a single framework.

    \section{Conclusion}\label{sec:conclusion}
    We consider the setting of fast dynamic view synthesis when only a few videos of the scene as observed from different static cameras are available.
    By exploiting the spatio-temporal correlation of the motion field, we design an explicit motion model using factorized representations that is compact, fast, and allows effective regularization with the flow priors.
    We observe that the naive use of cross-camera dense flow priors has a negative effect on the performance, while a careful imposition of reliable cross-camera sparse and within-camera dense flow priors provides a significant boost in performance.
    Further, we also show that the motion priors are more general and encompass the depth priors in dynamic view synthesis with sparse input views.
    We demonstrate the effectiveness of our approach on two popular datasets and show that our approach outperforms the state-of-the-art fast and compact dynamic radiance fields by a large margin when only a few viewpoints are available.

\begin{acks}
    This work was supported in part by a grant from Kotak IISc AI/ML centre (KIAC).
    The first author was supported partially by the Prime Minister’s Research Fellowship awarded by the Ministry of Education, Government of India.
\end{acks}

{
    \small
    \bibliographystyle{ACM-Reference-Format}
    \bibliography{DSSN}
}

    \input{tex/figures/qualitative_main_n3dv_3views}
    \input{tex/figures/qualitative_main_n3dv_4views}
    \input{tex/figures/qualitative_ablations_n3dv_3views}
    \input{tex/figures/qualitative_main_id_2views}
    \begin{figure*}
        \centering
        \begin{minipage}[b]{0.49\textwidth}
            \input{tex/figures/qualitative_depth_id_2views}
        \end{minipage}%
        \hfill
        \begin{minipage}[b]{0.49\textwidth}
            \input{tex/figures/qualitative_depth_n3dv_3views}
        \end{minipage}
    \end{figure*}
    \begin{figure*}
        \centering
        \begin{minipage}[b]{0.49\textwidth}
            \input{tex/figures/qualitative_naive_flow_n3dv_3views}
        \end{minipage}%
        \hfill
        \begin{minipage}[b]{0.49\textwidth}
            \input{tex/figures/quantitative_increasing_views_n3dv}
        \end{minipage}
    \end{figure*}

\appendix
\twocolumn[\subsection*{\centering \fontsize{13}{15}\selectfont Supplement}]

    \noindent The contents of this supplement include
    \begin{enumerate}[label=\Alph*., noitemsep]
        \item Video examples on N3DV and InterDigital datasets.
        \item Extensive quantitative evaluation reports.
    \end{enumerate}

    \section{Video Comparisons}\label{sec:video-comparisons}
    We compare our RF-DeRF model with K-Planes~\cite{fridovich2023kplanes}, ablated models, and various other baselines we design.
    We divide the video comparisons into three sets.
    In the first set, we compare our model against the vanilla K-Planes and the different baselines we design by incorporating sparse-depth and cross-camera dense flow priors.
    In the second set, we show the effect of removing each of our priors.
    Finally, we compare our base DeRF (without any priors) against K-Planes when dense input views are available.
    The videos are available on our project website \url{https://nagabhushansn95.github.io/publications/2024/RF-DeRF.html}.

    \section{Performance on Individual Scenes}\label{sec:performance-individual-scenes}
    For the benefit of follow-up work, where researchers may want to analyze the performance of different models or compare the models on individual scenes, we provide the performance of various models on individual scenes in \cref{tab:quantitative-scene-wise-n3dv-2views,tab:quantitative-scene-wise-n3dv-3views,tab:quantitative-scene-wise-n3dv-4views,tab:quantitative-scene-wise-id-2views,tab:quantitative-scene-wise-id-3views,tab:quantitative-scene-wise-id-4views}.

    \begin{table*}
        \centering
        \caption{Per-scene performance of various models with three input views on N3DV dataset.
        The four rows show PSNR, SSIM, LPIPS Alex, and Depth MAE scores, respectively.}
        \setlength{\tabcolsep}{3pt}
        \begin{tabular}{l|c|c|c|c|c|c|c}
            \hline
            \textbf{Model \textbackslash \ Scene Name} & \textbf{Coffee Martini} & \textbf{Cook Spinach} & \textbf{Cut Roasted Beef} & \textbf{Flame Salmon}  & \textbf{Flame Steak}  & \textbf{Sear Steak} & \textbf{Average} \\
            \hline

            HexPlane     & \makecell{13.70\\0.3892\\0.5814\\1.985} & \makecell{15.72\\0.5109\\0.4827\\2.0593} & \makecell{18.40\\0.6247\\0.3937\\1.6426} & \makecell{11.03\\0.3109\\0.6558\\2.3403} & \makecell{16.94\\0.5523\\0.451\\1.9577} & \makecell{17.42\\0.5705\\0.4296\\1.8327} & \makecell{15.53\\0.4931\\0.499\\1.9696} \\
            \hline
            K-Planes     & \makecell{20.11\\0.7590\\0.3215\\0.4501} & \makecell{24.24\\0.8743\\0.2059\\0.3275} & \makecell{26.55\\0.9128\\0.1767\\0.1570} & \makecell{16.92\\0.5975\\0.4717\\0.7002} & \makecell{26.77\\0.9178\\0.1541\\0.2178} & \makecell{27.29\\0.9216\\0.1520\\0.1718} & \makecell{23.65\\0.8305\\0.2470\\0.3374} \\
            \hline
            \makecell[l]{K-Planes + \\ Sparse Depth} & \makecell{20.66\\0.7830\\\textbf{0.2672}\\0.2910} & \makecell{22.70\\0.8557\\0.1916\\0.2468} & \makecell{26.66\\0.9090\\\textbf{0.1541}\\\textbf{0.1058}} & \makecell{17.19\\0.6167\\0.4149\\0.5459} & \makecell{24.52\\0.8767\\0.1700\\0.2011} & \makecell{25.02\\0.8971\\\textbf{0.1486}\\0.1640} & \makecell{22.79\\0.8230\\0.2244\\0.2591} \\
            \hline
            \makecell[l]{DeRF} & \makecell{19.82\\0.7531\\0.3282\\0.4619} & \makecell{23.15\\0.8361\\0.2338\\0.3320} & \makecell{25.19\\0.8946\\0.2018\\0.2080} & \makecell{16.93\\0.5852\\0.4579\\0.6320} & \makecell{25.33\\0.8819\\0.1917\\0.3483} & \makecell{25.25\\0.8805\\0.1909\\0.3340} & \makecell{22.61\\0.8052\\0.2674\\0.3860} \\
            \hline
            \makecell[l]{DeRF + \\ Sparse Depth} & \makecell{19.81\\0.7741\\0.3088\\0.3694} & \makecell{22.64\\0.8316\\0.2503\\0.3332} & \makecell{25.29\\0.8813\\0.2105\\0.1473} & \makecell{16.47\\0.5572\\0.4732\\0.6096} & \makecell{24.59\\0.8676\\0.2199\\0.3325} & \makecell{24.85\\0.8783\\0.1983\\0.2484} & \makecell{22.27\\0.7983\\0.2768\\0.3401} \\
            \hline
            \makecell[l]{DeRF + \\ Cross-Cam \\ Dense Flow} & \makecell{19.52\\0.6990\\0.4035\\0.6330} & \makecell{23.95\\0.8439\\0.2443\\0.4498} & \makecell{24.94\\0.8718\\0.2402\\0.3664} & \makecell{17.73\\0.6043\\0.4929\\0.7207} & \makecell{25.24\\0.8776\\0.2055\\0.3959} & \makecell{25.24\\0.8832\\0.1990\\0.3937} & \makecell{22.77\\0.7966\\0.2976\\0.4932} \\
            \hline
            \makecell[l]{RF-DeRF w/o \\ Sparse Flow} & \makecell{20.58\\0.8016\\0.2923\\0.3089} & \makecell{25.61\\\textbf{0.9052}\\\textbf{0.1808}\\0.2936} & \makecell{26.79\\0.9174\\0.1835\\0.2293} & \makecell{17.93\\0.6680\\0.3944\\0.5329} & \makecell{27.07\\0.9233\\\textbf{0.1533}\\0.2365} & \makecell{26.62\\0.9099\\0.1687\\0.2569} & \makecell{24.10\\0.8542\\0.2288\\0.3097} \\
            \hline
            \makecell[l]{RF-DeRF w/o \\ Dense Flow} & \makecell{21.10\\0.8106\\0.2863\\0.2648} & \makecell{26.03\\0.9003\\0.1875\\0.1791} & \makecell{\textbf{27.55}\\\textbf{0.9225}\\0.1760\\0.1254} & \makecell{19.56\\0.7452\\0.3622\\0.2866} & \makecell{26.85\\0.9153\\0.1671\\0.2060} & \makecell{\textbf{27.66}\\\textbf{0.9280}\\0.1518\\0.1457} & \makecell{24.79\\0.8703\\0.2218\\0.2013} \\
            \hline
            \makecell[l]{RF-DeRF} & \makecell{\textbf{21.63}\\\textbf{0.8233}\\0.2700\\\textbf{0.2095}} & \makecell{\textbf{26.29}\\0.9043\\0.1850\\\textbf{0.1601}} & \makecell{26.67\\0.9135\\0.1925\\0.1620} & \makecell{\textbf{20.77}\\\textbf{0.7776}\\\textbf{0.3194}\\\textbf{0.1877}} & \makecell{\textbf{27.95}\\\textbf{0.9330}\\0.1541\\\textbf{0.1320}} & \makecell{27.11\\0.9192\\0.1669\\\textbf{0.1345}} & \makecell{\textbf{25.07}\\\textbf{0.8785}\\\textbf{0.2146}\\\textbf{0.1643}} \\
            \hline

        \end{tabular}
        \label{tab:quantitative-scene-wise-n3dv-3views}
    \end{table*}

    \begin{table*}
        \centering
        \caption{Per-scene performance of various models with two input views on N3DV dataset.
        The four rows show PSNR, SSIM, LPIPS Alex, and Depth MAE scores, respectively.}
        \setlength{\tabcolsep}{3pt}
        \begin{tabular}{l|c|c|c|c|c|c|c}
            \hline
            \textbf{Model \textbackslash \ Scene Name} & \textbf{Coffee Martini} & \textbf{Cook Spinach} & \textbf{Cut Roasted Beef} & \textbf{Flame Salmon}  & \textbf{Flame Steak}  & \textbf{Sear Steak} & \textbf{Average} \\
            \hline

            HexPlane     & \makecell{13.10\\0.3984\\0.5565\\2.0923} & \makecell{14.14\\0.4284\\0.5848\\2.0748} & \makecell{15.16\\0.4507\\0.5165\\1.6882} & \makecell{10.56\\0.2518\\0.6963\\2.3864} & \makecell{15.04\\0.4735\\0.5673\\2.0209} & \makecell{14.84\\0.4707\\0.5504\\1.9361} & \makecell{13.81\\0.4123\\0.5786\\2.0331} \\
            \hline
            K-Planes     & \makecell{16.20\\0.5295\\0.4553\\0.7790} & \makecell{17.56\\0.5966\\0.4775\\0.5976} & \makecell{22.32\\0.8413\\0.2253\\0.2024} & \makecell{13.36\\0.3590\\0.6570\\0.9757} & \makecell{18.88\\0.6899\\0.3967\\0.4730} & \makecell{19.36\\0.7044\\0.3683\\0.4822} & \makecell{17.95\\0.6201\\0.4300\\0.5850} \\
            \hline
            \makecell[l]{K-Planes + \\ Sparse Depth} & \makecell{15.87\\0.5396\\0.4477\\0.7115} & \makecell{17.34\\0.6239\\0.4264\\0.4893} & \makecell{20.94\\0.8078\\0.2547\\0.2193} & \makecell{12.65\\0.3563\\0.6167\\\textbf{0.7521}} & \makecell{17.65\\0.6310\\0.4151\\0.5576} & \makecell{17.66\\0.6658\\0.3807\\0.4673} & \makecell{17.02\\0.6041\\0.4235\\0.5328} \\
            \hline
            \makecell[l]{DeRF} & \makecell{15.84\\0.5141\\0.4706\\0.8747} & \makecell{17.68\\0.6070\\0.4591\\0.6107} & \makecell{22.78\\0.8156\\0.2650\\0.3514} & \makecell{12.57\\0.3481\\0.6424\\1.0873} & \makecell{19.02\\0.6640\\0.4100\\0.5580} & \makecell{19.99\\0.7030\\0.3849\\0.5492} & \makecell{17.98\\0.6086\\0.4386\\0.6719} \\
            \hline
            \makecell[l]{RF-DeRF} & \makecell{\textbf{17.30}\\\textbf{0.6005}\\\textbf{0.4165}\\\textbf{0.5743}} & \makecell{\textbf{21.74}\\\textbf{0.7707}\\\textbf{0.3046}\\\textbf{0.3404}} & \makecell{\textbf{24.16}\\\textbf{0.8731}\\\textbf{0.1861}\\\textbf{0.1971}} & \makecell{\textbf{13.67}\\\textbf{0.4097}\\\textbf{0.6103}\\0.7844} & \makecell{\textbf{22.70}\\\textbf{0.8079}\\\textbf{0.2535}\\\textbf{0.3119}} & \makecell{\textbf{22.51}\\\textbf{0.8190}\\\textbf{0.2381}\\\textbf{0.2759}} & \makecell{\textbf{20.35}\\\textbf{0.7135}\\\textbf{0.3349}\\\textbf{0.4140}} \\
            \hline

        \end{tabular}
        \label{tab:quantitative-scene-wise-n3dv-2views}
    \end{table*}

    \begin{table*}
        \centering
        \caption{Per-scene performance of various models with four input views on N3DV dataset.
        The four rows show PSNR, SSIM, LPIPS Alex, and Depth MAE scores, respectively.}
        \setlength{\tabcolsep}{3pt}
        \begin{tabular}{l|c|c|c|c|c|c|c}
            \hline
            \textbf{Model \textbackslash \ Scene Name} & \textbf{Coffee Martini} & \textbf{Cook Spinach} & \textbf{Cut Roasted Beef} & \textbf{Flame Salmon}  & \textbf{Flame Steak}  & \textbf{Sear Steak} & \textbf{Average} \\
            \hline

            HexPlane & \makecell{13.99\\0.3846\\0.5682\\1.9928} & \makecell{14.52\\0.4299\\0.5939\\2.0695} & \makecell{21.92\\0.8078\\0.2289\\1.3548} & \makecell{11.21\\0.316\\0.6671\\2.3662} & \makecell{15.76\\0.4482\\0.5837\\2.0197} & \makecell{15.62\\0.4429\\0.5701\\1.9188} & \makecell{15.50\\0.4716\\0.5353\\1.9536} \\
            \hline
            K-Planes     & \makecell{22.89\\0.8603\\\textbf{0.1971}\\0.3168} & \makecell{24.71\\\textbf{0.8702}\\\textbf{0.1911}\\0.2588} & \makecell{\textbf{28.64}\\\textbf{0.9418}\\\textbf{0.1068}\\0.1506} & \makecell{18.91\\0.6985\\0.3657\\0.5428} & \makecell{26.90\\\textbf{0.9040}\\\textbf{0.1632}\\\textbf{0.1578}} & \makecell{26.73\\0.9051\\\textbf{0.1510}\\0.1730} & \makecell{24.80\\0.8633\\0.1958\\0.2666} \\
            \hline
            \makecell[l]{K-Planes + \\ Sparse Depth} & \makecell{20.26\\0.7938\\0.2646\\0.2291} & \makecell{21.08\\0.7696\\0.2957\\0.3044} & \makecell{25.69\\0.9029\\0.1607\\\textbf{0.1493}} & \makecell{18.14\\0.6907\\0.3608\\0.2761} & \makecell{21.28\\0.7564\\0.2959\\0.3128} & \makecell{22.47\\0.8052\\0.2454\\0.2662} & \makecell{21.49\\0.7864\\0.2705\\0.2563} \\
            \hline
            \makecell[l]{DeRF} & \makecell{22.02\\0.8468\\0.2290\\0.3545} & \makecell{23.36\\0.8167\\0.2603\\0.3749} & \makecell{27.91\\0.9295\\0.1306\\0.1590} & \makecell{17.70\\0.6273\\0.4303\\0.6675} & \makecell{25.72\\0.8679\\0.2069\\0.3359} & \makecell{25.86\\0.8857\\0.1794\\0.2631} & \makecell{23.76\\0.8290\\0.2394\\0.3591} \\
            \hline
            \makecell[l]{RF-DeRF} & \makecell{\textbf{23.43}\\\textbf{0.8798}\\0.2089\\\textbf{0.1536}} & \makecell{\textbf{25.76}\\0.8699\\0.2101\\\textbf{0.2093}} & \makecell{28.32\\0.9310\\0.1364\\0.1523} & \makecell{\textbf{22.85}\\\textbf{0.8629}\\\textbf{0.2329}\\\textbf{0.1524}} & \makecell{\textbf{27.10}\\0.8979\\0.1722\\0.2077} & \makecell{\textbf{27.68}\\\textbf{0.9070}\\0.1621\\\textbf{0.1685}} & \makecell{\textbf{25.86}\\\textbf{0.8914}\\\textbf{0.1871}\\\textbf{0.1740}} \\
            \hline

        \end{tabular}
        \label{tab:quantitative-scene-wise-n3dv-4views}
    \end{table*}

    \begin{table*}
        \centering
        \caption{Per-scene performance of various models with three input views on InterDigital dataset.
        The four rows show PSNR, SSIM, LPIPS Alex, and Depth MAE scores, respectively.}
        \begin{tabular}{l|c|c|c|c|c|c}
            \hline
            \textbf{Model \textbackslash \ Scene Name} & \textbf{Birthday} & \textbf{Painter} & \textbf{Remy} & \textbf{Theater}  & \textbf{Train}  & \textbf{Average} \\
            \hline

            HexPlane & \makecell{12.54\\0.2115\\0.5298\\1.2777} & \makecell{17.42\\0.38\\0.433\\1.4585} & \makecell{13.34\\0.4295\\0.4888\\1.1883} & \makecell{14.75\\0.3153\\0.5735\\1.3894} & \makecell{11.24\\0.0001\\0.6867\\1.8348} & \makecell{13.86\\0.2673\\0.5424\\1.4297} \\
            \hline
            K-Planes     & \makecell{19.58\\0.8639\\\textbf{0.1388}\\0.0886} & \makecell{26.42\\0.9369\\\textbf{0.0892}\\\textbf{0.0700}} & \makecell{15.18\\0.4302\\0.5086\\0.2936} & \makecell{17.01\\0.5339\\0.3949\\0.2530} & \makecell{15.56\\0.5591\\0.3918\\0.3846} & \makecell{18.75\\0.6648\\0.3047\\0.2179} \\
            \hline
            \makecell[l]{K-Planes + \\ Sparse Depth} & \makecell{19.49\\0.8476\\0.1525\\0.1194} & \makecell{26.26\\0.9305\\0.0977\\0.1708} & \makecell{15.59\\0.4929\\0.4580\\0.2091} & \makecell{17.59\\0.5744\\0.3737\\0.3036} & \makecell{17.84\\0.6870\\0.2792\\0.2189} & \makecell{19.36\\0.7065\\0.2722\\0.2044} \\
            \hline
            \makecell[l]{DeRF} & \makecell{19.55\\0.8636\\0.1533\\0.0987} & \makecell{25.93\\0.9272\\0.1037\\0.1344} & \makecell{16.39\\0.5377\\0.4413\\0.2121} & \makecell{16.69\\0.5248\\0.4144\\0.2608} & \makecell{15.02\\0.5443\\0.3990\\0.3938} & \makecell{18.72\\0.6795\\0.3024\\0.2200} \\
            \hline
            \makecell[l]{DeRF + \\ Sparse Depth} & \makecell{19.25\\0.8441\\0.1615\\0.1149} & \makecell{25.98\\0.9249\\0.1107\\0.2323} & \makecell{15.34\\0.5037\\0.4566\\0.2335} & \makecell{16.02\\0.5338\\0.4081\\0.2832} & \makecell{18.03\\0.6691\\0.2908\\0.2155} & \makecell{18.92\\0.6951\\0.2855\\0.2159} \\
            \hline
            \makecell[l]{DeRF + \\ Cross-Cam \\ Dense Flow} & \makecell{18.25\\0.7215\\0.3034\\0.2503} & \makecell{24.56\\0.8613\\0.1802\\0.3576} & \makecell{16.20\\0.5050\\0.4735\\0.3198} & \makecell{16.10\\0.4545\\0.4761\\0.2397} & \makecell{17.13\\0.6095\\0.3677\\0.2720} & \makecell{18.45\\0.6304\\0.3602\\0.2879} \\
            \hline
            \makecell[l]{RF-DeRF w/o \\ Sparse Flow} & \makecell{19.26\\0.8339\\0.1946\\0.1624} & \makecell{25.91\\0.9099\\0.1245\\0.3070} & \makecell{15.73\\0.4918\\0.4974\\0.2781} & \makecell{17.23\\0.5346\\0.4108\\0.2833} & \makecell{17.27\\0.6556\\0.3138\\0.2126} & \makecell{19.08\\0.6851\\0.3082\\0.2487} \\
            \hline
            \makecell[l]{RF-DeRF w/o \\ Dense Flow} & \makecell{19.51\\\textbf{0.8723}\\0.1412\\\textbf{0.0666}} & \makecell{26.51\\\textbf{0.9397}\\0.0917\\0.0701} & \makecell{\textbf{17.69}\\0.5594\\0.4154\\0.2036} & \makecell{18.45\\0.5967\\\textbf{0.3436}\\0.2112} & \makecell{18.14\\0.6901\\0.2874\\0.0775} & \makecell{20.06\\0.7316\\0.2559\\0.1258} \\
            \hline
            \makecell[l]{RF-DeRF} & \makecell{\textbf{19.91}\\0.8680\\0.1449\\0.0718} & \makecell{\textbf{26.55}\\0.9306\\0.1038\\0.0953} & \makecell{17.51\\\textbf{0.5749}\\\textbf{0.4036}\\\textbf{0.1929}} & \makecell{\textbf{18.84}\\\textbf{0.6092}\\0.3471\\\textbf{0.1952}} & \makecell{\textbf{19.24}\\\textbf{0.7265}\\\textbf{0.2546}\\\textbf{0.0639}} & \makecell{\textbf{20.41}\\\textbf{0.7418}\\\textbf{0.2508}\\\textbf{0.1238}} \\
            \hline

        \end{tabular}
        \label{tab:quantitative-scene-wise-id-3views}
    \end{table*}

    \bigskip

    \begin{table*}
        \centering
        \caption{Per-scene performance of various models with two input views on InterDigital dataset.
        The four rows show PSNR, SSIM, LPIPS Alex, and Depth MAE scores, respectively.}
        \begin{tabular}{l|c|c|c|c|c|c}
            \hline
            \textbf{Model \textbackslash \ Scene Name} & \textbf{Birthday} & \textbf{Painter} & \textbf{Remy} & \textbf{Theater}  & \textbf{Train}  & \textbf{Average} \\
            \hline

            HexPlane     & \makecell{11.94\\0.1623\\0.578\\1.3664} & \makecell{16.58\\0.2151\\0.4826\\1.6554} & \makecell{13.75\\0.4065\\0.4829\\1.2256} & \makecell{14.32\\0.2497\\0.6028\\1.4564} & \makecell{12.31\\0.1499\\0.6172\\1.8696} & \makecell{13.78\\0.2367\\0.5527\\1.5147} \\
            \hline
            K-Planes     & \makecell{16.71\\0.6798\\0.2726\\0.2105} & \makecell{20.23\\0.6687\\0.2905\\0.3262} & \makecell{14.18\\0.4016\\0.5047\\0.3790} & \makecell{16.24\\0.4168\\0.4573\\0.3061} & \makecell{15.20\\0.3991\\0.4852\\0.5471} & \makecell{16.51\\0.5132\\0.4021\\0.3538} \\
            \hline
            \makecell[l]{DeRF} & \makecell{17.76\\0.7572\\0.2286\\0.1782} & \makecell{21.27\\0.7271\\0.2487\\0.3787} & \makecell{15.00\\0.4778\\\textbf{0.4659}\\\textbf{0.3275}} & \makecell{16.12\\0.4332\\0.4600\\0.3477} & \makecell{15.06\\0.3958\\0.4590\\0.5409} & \makecell{17.04\\0.5582\\0.3725\\0.3546} \\
            \hline
            \makecell[l]{RF-DeRF} & \makecell{\textbf{19.08}\\\textbf{0.7913}\\\textbf{0.2239}\\\textbf{0.1688}} & \makecell{\textbf{24.38}\\\textbf{0.8360}\\\textbf{0.1855}\\\textbf{0.2542}} & \makecell{\textbf{15.62}\\\textbf{0.4952}\\0.4728\\0.3404} & \makecell{\textbf{17.68}\\\textbf{0.4905}\\\textbf{0.4050}\\\textbf{0.2870}} & \makecell{\textbf{18.64}\\\textbf{0.6914}\\\textbf{0.2837}\\\textbf{0.1124}} & \makecell{\textbf{19.08}\\\textbf{0.6609}\\\textbf{0.3142}\\\textbf{0.2326}} \\
            \hline

        \end{tabular}
        \label{tab:quantitative-scene-wise-id-2views}
    \end{table*}

    \begin{table*}
        \centering
        \caption{Per-scene performance of various models with four input views on InterDigital dataset.
        The four rows show PSNR, SSIM, LPIPS Alex, and Depth MAE scores, respectively.}
        \begin{tabular}{l|c|c|c|c|c|c}
            \hline
            \textbf{Model \textbackslash \ Scene Name} & \textbf{Birthday} & \textbf{Painter} & \textbf{Remy} & \textbf{Theater}  & \textbf{Train}  & \textbf{Average} \\
            \hline

            K-Planes     & \makecell{23.26\\\textbf{0.9294}\\\textbf{0.0895}\\0.0553} & \makecell{\textbf{30.33}\\\textbf{0.9708}\\\textbf{0.0642}\\\textbf{0.0547}} & \makecell{21.04\\0.7512\\0.2747\\0.1549} & \makecell{\textbf{23.72}\\\textbf{0.8288}\\\textbf{0.1897}\\\textbf{0.1475}} & \makecell{20.93\\0.8163\\0.1674\\0.1448} & \makecell{\textbf{23.85}\\\textbf{0.8593}\\\textbf{0.1571}\\0.1114} \\
            \hline
            \makecell[l]{DeRF} & \makecell{\textbf{23.31}\\0.9243\\0.1013\\0.0697} & \makecell{29.60\\0.9662\\0.0725\\0.0965} & \makecell{21.46\\\textbf{0.7649}\\\textbf{0.2552}\\0.1431} & \makecell{22.45\\0.7598\\0.2444\\0.1980} & \makecell{20.72\\0.8114\\0.1879\\0.1385} & \makecell{23.51\\0.8453\\0.1723\\0.1292} \\
            \hline
            \makecell[l]{RF-DeRF} & \makecell{23.02\\0.9228\\0.0992\\\textbf{0.0519}} & \makecell{29.22\\0.9671\\0.0688\\0.0706} & \makecell{\textbf{21.74}\\0.7571\\0.2606\\\textbf{0.1398}} & \makecell{22.82\\0.7748\\0.2484\\0.1776} & \makecell{\textbf{21.65}\\\textbf{0.8546}\\\textbf{0.1541}\\\textbf{0.0632}} & \makecell{23.69\\0.8553\\0.1662\\\textbf{0.1006}} \\
            \hline

        \end{tabular}
        \label{tab:quantitative-scene-wise-id-4views}
    \end{table*}
\end{document}

%% file: tex/tables/related_work.tex
\begin{table}
    \centering
    \setlength{\tabcolsep}{4pt}
    \begin{tabular}{l|cccc}
        \hline
        & \makecell{Dynamic \\ Scenes} & \makecell{Sparse \\ Views} & \makecell{Explicit \\ Model} & \makecell{Motion \\ Model} \\
        \hline
        NeRF & \xmark & \xmark & \xmark & -- \\
        TensoRF, i-ngp, 3DGS & \xmark & \xmark & \cmark & -- \\
        DS-NeRF, SimpleNeRF & \xmark & \cmark & \xmark & -- \\
        NSFF$^*$, DyNeRF & \cmark & \xmark & \xmark & \xmark \\
        D-NeRF & \cmark & \xmark & \xmark & \cmark \\
        TiNeuVox, SWAGS & \cmark & \xmark & \ \xmark$^\dagger$ & \cmark \\
        K-Planes, HexPlane & \cmark & \xmark & \cmark & \xmark \\
        Ours & \cmark & \cmark & \cmark & \cmark \\
        \hline
    \end{tabular}
    \caption{
        \textbf{Related work overview:} We compare our work with prior works based on various aspects. Sparse input views refers to models that can handle data from few stationary multi-view cameras. Explicit refers to models that primarily employ explicit models followed by an optional tiny MLP.\\
        $^\dagger$ TiNeuVox and SWAGS use an explicit model to represent the scene, but use implicit neural networks to model the motion. $^*$ NSFF predicts motion as an auxiliary task.
    }
    \label{tab:related-work}
\end{table}

%% file: tex/figures/model_architecture.tex
\begin{figure*}
    \centering
    \includegraphics[width=\linewidth]{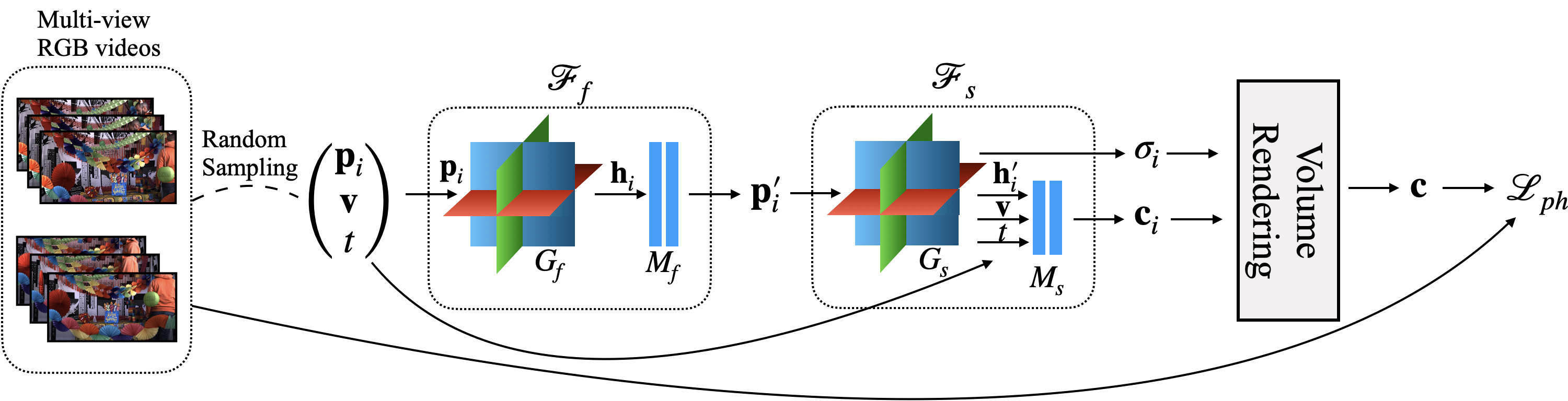}
    \caption{
        \textbf{Model architecture:}
        We decompose the dynamic radiance field into a 4D scene flow or deformation field $\tensorfflow$ that maps a 3D point $\pointp_i$ at time $t$ to the corresponding 3D point $\pointp_i'$ at canonical time $t'$, and a 5D radiance field $\tensorfscene$ that models the scene at canonical time $t'$.
        Both the fields are modeled using a factorized volume followed by a tiny MLP, which allows fast optimization and rendering.
        We note that $\gridg_f$ is modeled using six planes, although we show only three owing to the difficulty in visualizing four dimensions.
        The MLP $\mlpm_s$ is conditioned on time and viewing direction to model time-dependent color variations such as shadows and view-dependent color variations such as specularities.
        The output of $\tensorfscene$ is volume rendered to obtain the color of the pixel and the photometric loss is used to train both the fields.
        The explicitly modeled motion field $\tensorfflow$ is additionally regularized using the flow priors as shown in \cref{fig:loss-flow}.
    }
    \label{fig:model-architecture}
\end{figure*}

%% file: tex/figures/loss_flow.tex
\begin{figure}
    \centering
    \includegraphics[width=\linewidth]{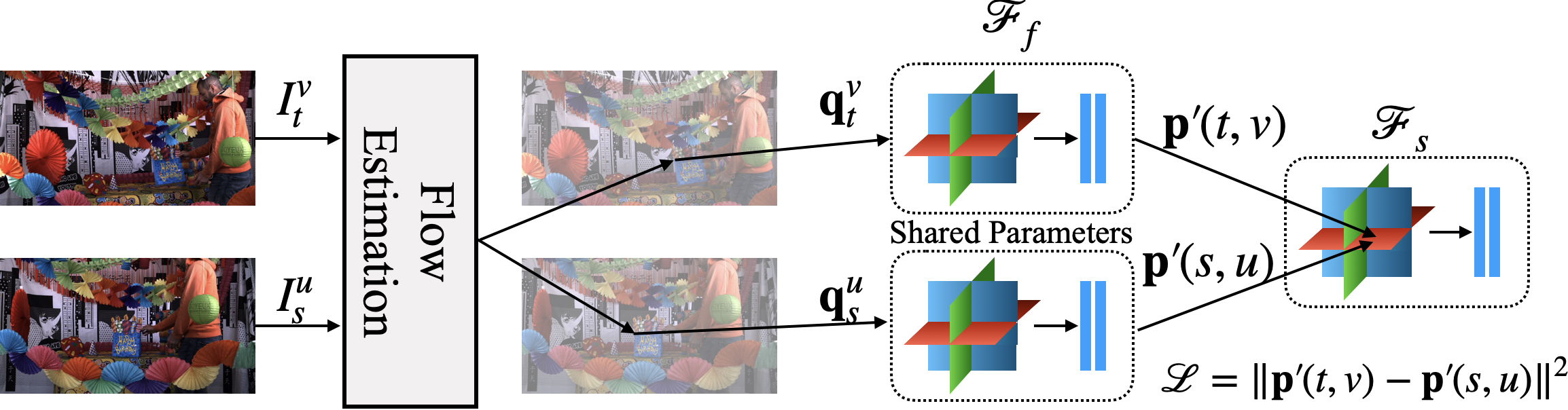}
    \caption{
        \textbf{Flow regularization:}
        Since the motion field $\tensorfflow$ gives only the unidirectional flow from time $t$ to $t'$, we impose the flow prior by minimizing the distance between the 3D points in the canonical volume corresponding to the matched pixels $(\pixelq_t^v, \pixelq_s^u)$ in the input frames $(I_t^v,I_s^u)$.
    }
    \label{fig:loss-flow}
\end{figure}

%% file: tex/figures/flow_matches.tex

\begin{figure}
    \centering
    \begin{subfigure}[b]{0.9\linewidth}
        \centering
        \includegraphics[width=\linewidth]{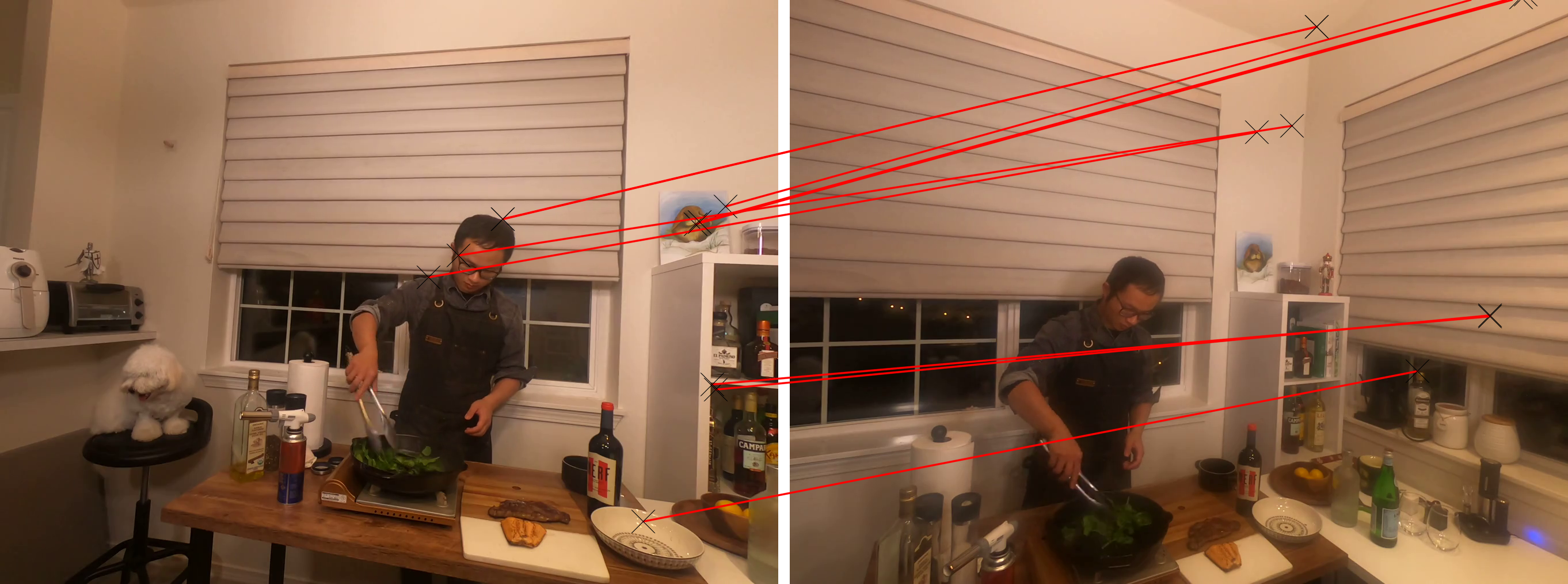}
        \caption{
               Dense flow across cameras estimated by RAFT.
        }
        \label{fig:cross-view-dense-flow}
    \end{subfigure}
    \begin{subfigure}[b]{0.9\linewidth}
        \centering
        \includegraphics[width=\linewidth]{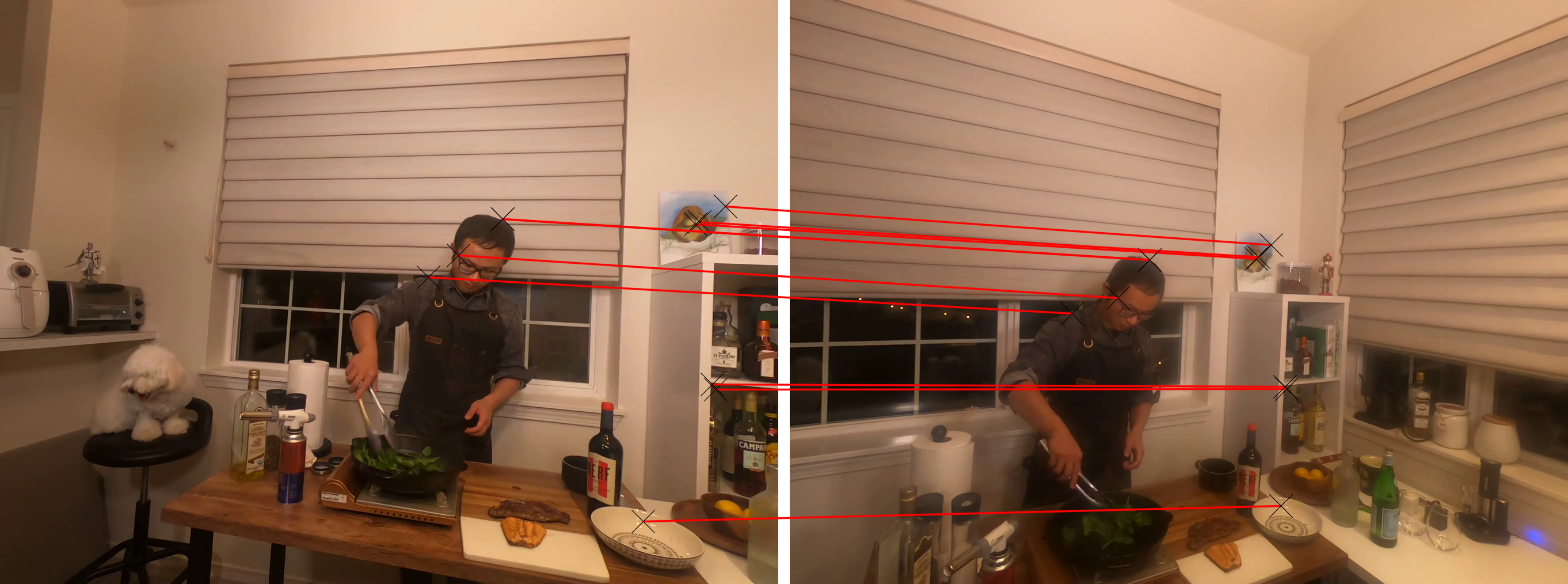}
        \caption{
               Sparse flow across cameras estimated by SIFT.
        }
        \label{fig:cross-view-sparse-flow}
    \end{subfigure}
    \begin{subfigure}[b]{0.9\linewidth}
        \centering
        \includegraphics[width=\linewidth]{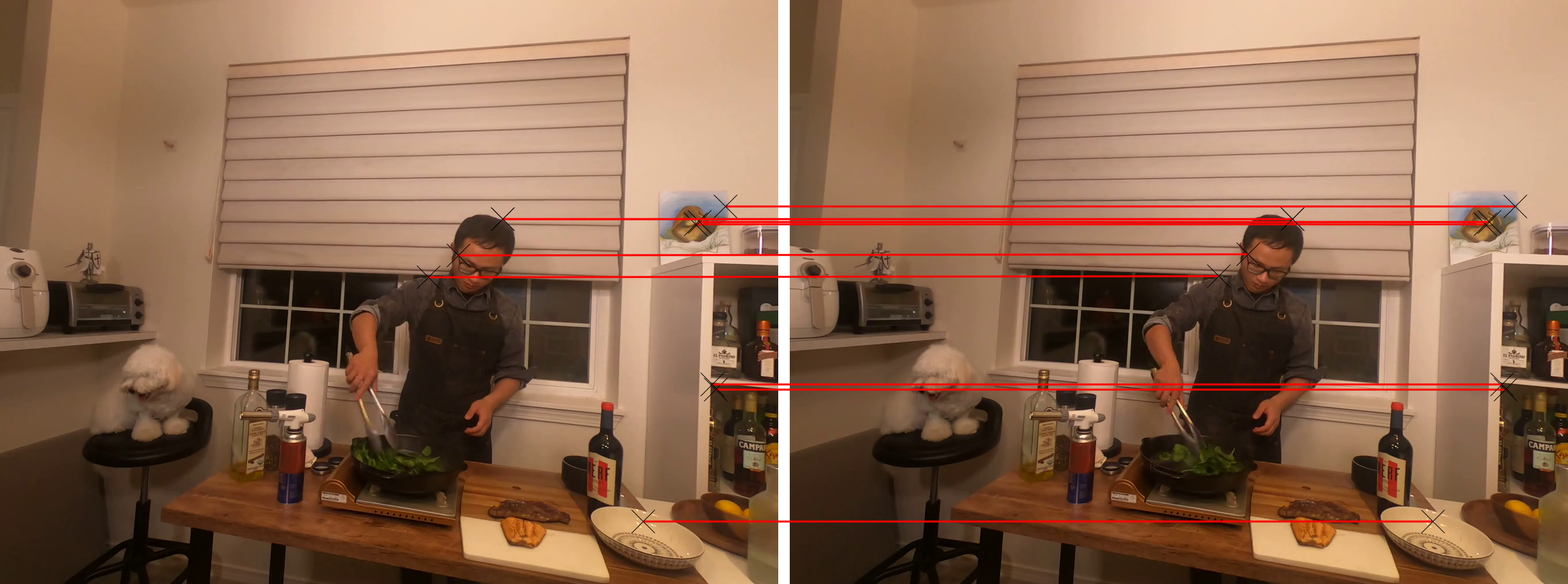}
        \caption{
               Dense flow within cameras estimated by RAFT.
        }
        \label{fig:same-view-dense-flow}
    \end{subfigure}
    \caption{
        \textbf{Visualization of different flow priors:}
        We show the matched pixels as provided by different flow priors.
        The pixels in the first view are randomly picked from those for which sparse flow is available and the same pixels are used for dense flow.
        Note that the second view in the first two examples has more blur as compared to the first view.
        (a) We show that the dense flow priors across cameras obtained using deep optical flow networks such as RAFT~\cite{teed2020raft} are prone to erroneous matches, due to variations in camera parameters and lighting.
        We observed similar trends with other deep optical flow networks as well such as AR-Flow~\cite{liu2020arflow}.
        (b) Matching pixels across cameras using robust SIFT features provides reliable matches, albeit sparse.
        (c) Within individual cameras, the dense correspondences provided by deep optical flow networks are more reliable owing to smaller variations in lighting.
    }
    \label{fig:failure-dense-flow}
\end{figure}

%% file: tex/tables/quantitative_main.tex
    \begin{table*}
        \centering
            \begin{tabular}{c|l|cccc|cccc}
                \hline
                & & \multicolumn{4}{c|}{\textbf{N3DV}} & \multicolumn{4}{c}{\textbf{InterDigital}} \\
                & Model & \makecell{PSNR $\uparrow$} & \makecell{SSIM $\uparrow$} & \makecell{LPIPS $\downarrow$} & \makecell{Depth MAE $\downarrow$} & \makecell{PSNR $\uparrow$} & \makecell{SSIM $\uparrow$} & \makecell{LPIPS $\downarrow$} & \makecell{Depth MAE $\downarrow$} \\
                \hline
                \parbox[t]{2mm}{\multirow{3}{*}{\rotatebox[origin=c]{90}{2 views}}}
                & HexPlane & 13.81 & 0.41 & 0.58 & 2.03 & 13.78 & 0.24 & 0.55 & 1.51 \\  
                & K-Planes & 17.95 & 0.62 & 0.43 & 0.59 & 16.51 & 0.51 & 0.40 & 0.35 \\  
                & DeRF & 17.98 & 0.61 & 0.44 & 0.67 & 17.04 & 0.56 & 0.37 & 0.35 \\  
                & RF-DeRF & \textbf{20.35} & \textbf{0.71} & \textbf{0.33} & \textbf{0.41} & \textbf{19.08} & \textbf{0.66} & \textbf{0.31} & \textbf{0.23} \\  
                \hline
                \parbox[t]{2mm}{\multirow{3}{*}{\rotatebox[origin=c]{90}{3 views}}}
                & HexPlane & 15.53 & 0.49 & 0.50 & 1.97 & 13.85 & 0.27 & 0.54 & 1.43 \\  
                & K-Planes & 23.65 & 0.83 & 0.25 & 0.34 & 18.75 & 0.66 & 0.30 & 0.22 \\  
                & DeRF & 22.61 & 0.81 & 0.27 & 0.39 & 18.72 & 0.68 & 0.30 & 0.22 \\  
                & RF-DeRF & \textbf{25.07} & \textbf{0.88} & \textbf{0.21} & \textbf{0.16} & \textbf{20.41} & \textbf{0.74} & \textbf{0.25} & \textbf{0.12} \\  
                \hline
                \parbox[t]{2mm}{\multirow{3}{*}{\rotatebox[origin=c]{90}{4 views}}}
                & HexPlane & 15.50 & 0.47 & 0.54 & 1.95 & 15.18 & 0.38 & 0.49 & 1.24 \\  
                & K-Planes & 24.80 & 0.86 & 0.20 & 0.27 & \textbf{23.85} & \textbf{0.86} & \textbf{0.16} & 0.11 \\  
                & DeRF & 23.76 & 0.83 & 0.24 & 0.36 & 23.51 & 0.85 & 0.17 & 0.13 \\  
                & RF-DeRF & \textbf{25.86} & \textbf{0.89} & \textbf{0.19} & \textbf{0.17} & 23.69 & \textbf{0.86} & 0.17 & \textbf{0.10} \\ 
                \hline
            \end{tabular}
        \caption{
            \textbf{Quantitative results:}
            We compare our model with K-Planes on N3DV~\cite{li2022neural} and InterDigital~\cite{sabater2017interdigital} datasets with two, three, and four input views.
            We also show the performance of our base DeRF model for reference.
            We report the PSNR, SSIM, and LPIPS scores for the rendered images and the depth MAE for the rendered depth maps.
            The best scores in each category are shown in bold.
        }
        \label{tab:quantitative-results-main}
    \end{table*}

%% file: tex/figures/qualitative_main_n3dv_2views.tex
\begin{figure}
    \centering
    \includegraphics[width=\linewidth]{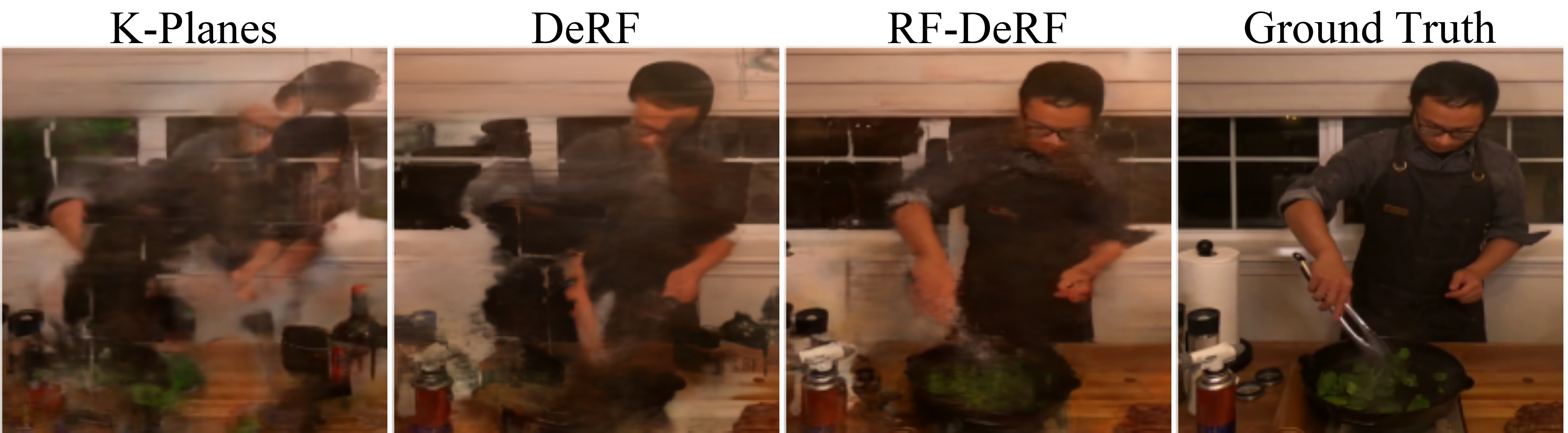}
    \caption{
        \textbf{Qualitative examples on N3DV dataset with two input views:}
        We can observe that K-Planes finds it hard to learn the moving person leading to significant distortions.
        Our DeRF model (without any priors) corrects a few errors by virtue of the common canonical volume.
        Imposing our priors leads to much better reconstruction.
    }
    \label{fig:qualitative-main-n3dv-2views}
\end{figure}

%% file: tex/tables/quantitative_ablations.tex
\begin{table}
    \centering
    \begin{tabular}{cc|cccc}
        \hline
        $\sparseFlowPrior$ & $\denseFlowPrior$ & \makecell{PSNR $\uparrow$} & \makecell{SSIM $\uparrow$} & \makecell{LPIPS $\downarrow$} & \makecell{Depth MAE $\downarrow$} \\
        \hline
        & & 22.61 & 0.81 & 0.27 & 0.39 \\  
        \cmark & & 24.79 & 0.87 & 0.22 & 0.20 \\  
        & \cmark & 24.10 & 0.85 & 0.23 & 0.31 \\  
        \cmark & \cmark & \textbf{25.07} & \textbf{0.88} & \textbf{0.21} & \textbf{0.16} \\  
        \hline
    \end{tabular}
    \caption{
        \textbf{Quantitative results of ablations on N3DV dataset with three input views:}
        We evaluate the significance of the two flow priors we employ, namely the sparse flow prior ($\sparseFlowPrior$) and the dense flow prior ($\denseFlowPrior$) by disabling one component at a time.
        We also show the performance of the DeRF model without any priors for reference.
    }
    \label{tab:quantitative-results-ablations}
\end{table}

%% file: tex/tables/quantitative_cross_camera_dense_flow.tex
\begin{table}
    \centering
    \setlength{\tabcolsep}{3pt}
    \begin{tabular}{l|cc|cc}
        \hline
        & \multicolumn{2}{c|}{\textbf{N3DV}} & \multicolumn{2}{c}{\textbf{InterDigital}} \\
        Model & LPIPS $\downarrow$ & \makecell{Depth \\ MAE} $\downarrow$ & LPIPS $\downarrow$ & \makecell{Depth \\ MAE} $\downarrow$ \\
        \hline
        w/o any priors & 0.27 & 0.39 & 0.30 & 0.22 \\  
        w/ cross-camera dense flow & 0.30 & 0.49 & 0.36 & 0.29 \\  
        w/ our flow priors & \textbf{0.21} & \textbf{0.16} & \textbf{0.25} & \textbf{0.12} \\  
        \hline
    \end{tabular}
    \caption{
    We test the performance of our DeRF model with dense flow priors across cameras and our reliable flow priors.
    The performance of the DeRF model (without any priors) is also shown for reference.
    }
    \label{tab:cross-camera-dense-flow}
\end{table}

%% file: tex/figures/qualitative_sparse_depth_n3dv_3views.tex
\begin{figure*}
    \centering
    \includegraphics[width=\linewidth]{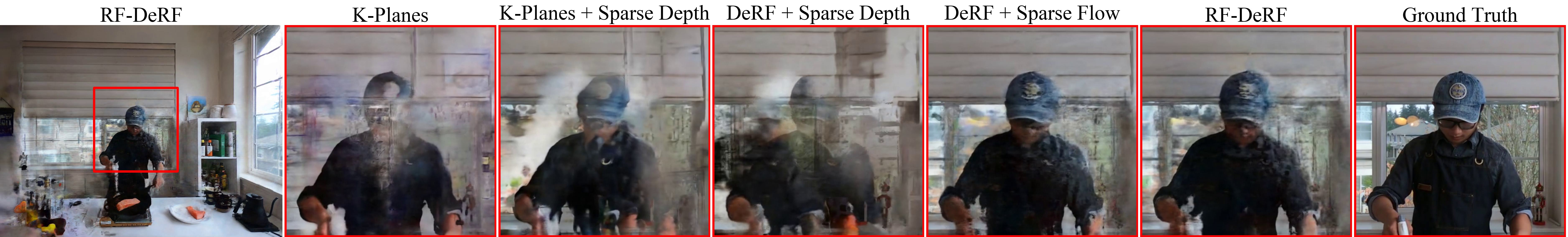}
    \caption{
        \textbf{Qualitative examples for depth priors vs flow priors:}
        We observe that sparse depth prior is not very effective in improving the reconstruction quality.
        However, our sparse flow prior is highly effective in mitigating the distortions, while the use of sparse and dense flow priors in our final model leads to the best reconstruction quality.
    }
    \label{fig:qualitative-sparse-depth-n3dv-3views}
\end{figure*}

%% file: tex/tables/quantitative_sparse_depth.tex
\begin{table}
    \centering
    \setlength{\tabcolsep}{3pt}
    \begin{tabular}{l|cc|cc}
        \hline
        & \multicolumn{2}{c|}{\textbf{N3DV}} & \multicolumn{2}{c}{\textbf{InterDigital}} \\
        Model & SSIM $\uparrow$ & \makecell{Depth MAE} $\downarrow$ & SSIM $\uparrow$ & \makecell{Depth MAE} $\downarrow$ \\
        \hline
        K-Planes & 0.83 & 0.34 & 0.66 & 0.22 \\  
        K-Planes + $\sparseDepthPrior$ & 0.82 & 0.26 & 0.71 & 0.20 \\  
        \hline
        DeRF & 0.81 & 0.39 & 0.68 & 0.22 \\  
        DeRF + $\sparseDepthPrior$ & 0.80 & 0.34 & 0.70 & 0.22 \\  
        DeRF + $\sparseFlowPrior$ & 0.87 & 0.20 & 0.73 & 0.13 \\  
        RF-DeRF & \textbf{0.88} & \textbf{0.16} & \textbf{0.74} & \textbf{0.12} \\  
        \hline
    \end{tabular}
    \caption{
    We analyze the impact of our sparse flow prior that regularizes both the scene geometry and the learned motion against the sparse depth prior~\cite{deng2022dsnerf} that regularizes only the scene geometry.
    The models are trained with three input views on both the datasets.
    }
    \label{tab:sparse-depth-vs-sparse-flow}
\end{table}

%% file: tex/tables/quantitative_dense_views.tex
\begin{table}
    \centering
    \setlength{\tabcolsep}{3pt}
    \begin{tabular}{l|ccc|ccc}
        \hline
        & \multicolumn{3}{c|}{\textbf{N3DV}} & \multicolumn{3}{c}{\textbf{InterDigital}} \\
        Model & PSNR $\uparrow$ & SSIM $\uparrow$ & LPIPS $\downarrow$ & PSNR $\uparrow$ & SSIM $\uparrow$ & LPIPS $\downarrow$ \\
        \hline
        K-Planes & 30.55 & 0.96 & 0.12 & 29.00 & 0.96 & 0.07 \\  
        DeRF & 29.95 & 0.95 & 0.12 & 28.14 & 0.95 & 0.09 \\  
        \hline
    \end{tabular}
    \caption{
    We test the performance of our DeRF model against K-Planes with dense input views.
    Please see the videos in the supplementary for visual comparisons.
    }
    \label{tab:dense-input-views}
\end{table}

%% file: tex/figures/qualitative_main_n3dv_3views.tex
\begin{figure*}
    \centering
    \includegraphics[width=0.85\linewidth]{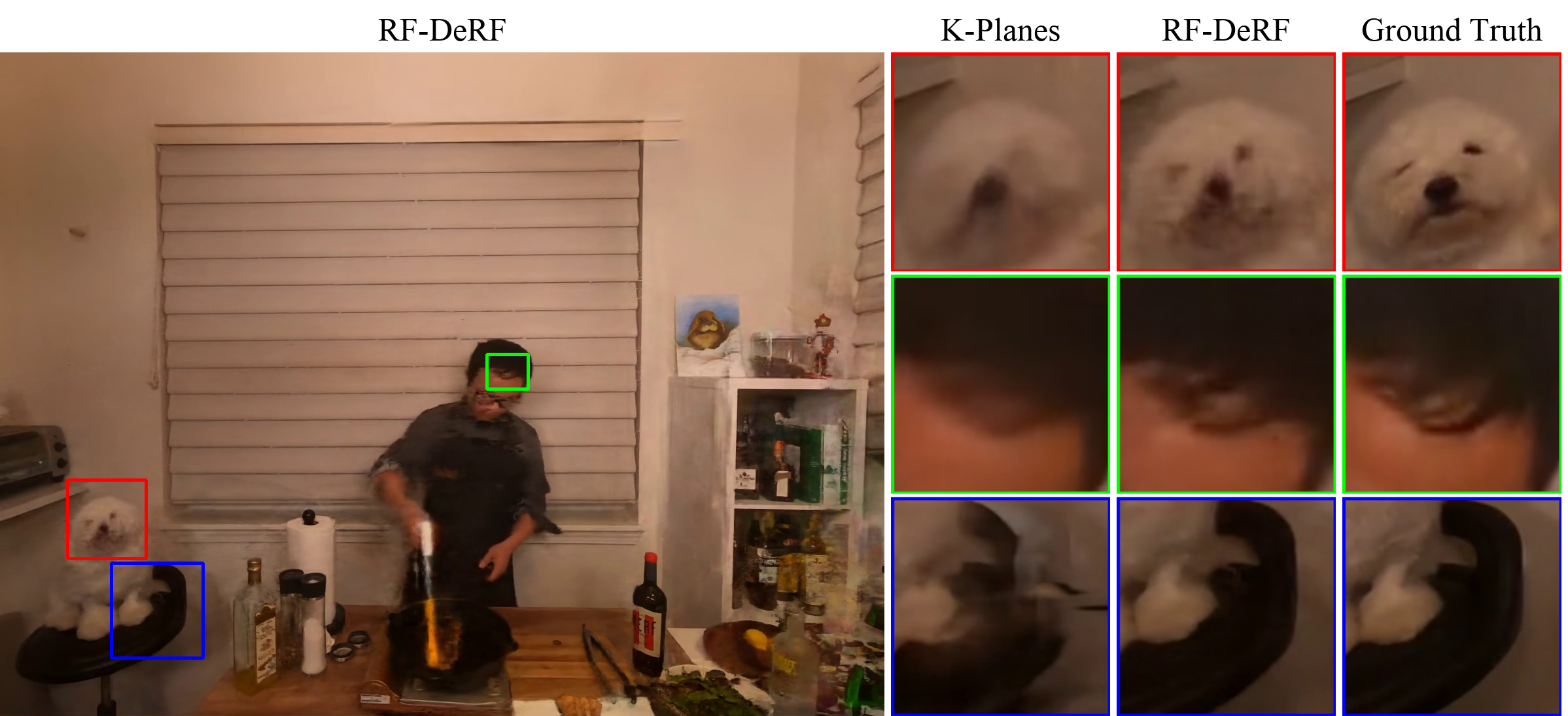}
    \caption{
        \textbf{Qualitative examples on N3DV dataset with three input views:}
        We observe that K-Planes blurs moving objects such as the face of the dog and the hairs on the face of the person in the first two examples respectively.
        We also find that K-Planes creates distortions in the static regions when an object moves in its vicinity as seen in the third row.
        However, our RF-DeRF model produces sharper and more accurate results.
    }
    \label{fig:qualitative-main-n3dv-3views}
\end{figure*}

%% file: tex/figures/qualitative_main_n3dv_4views.tex
\begin{figure*}
    \centering
    \includegraphics[width=0.85\linewidth]{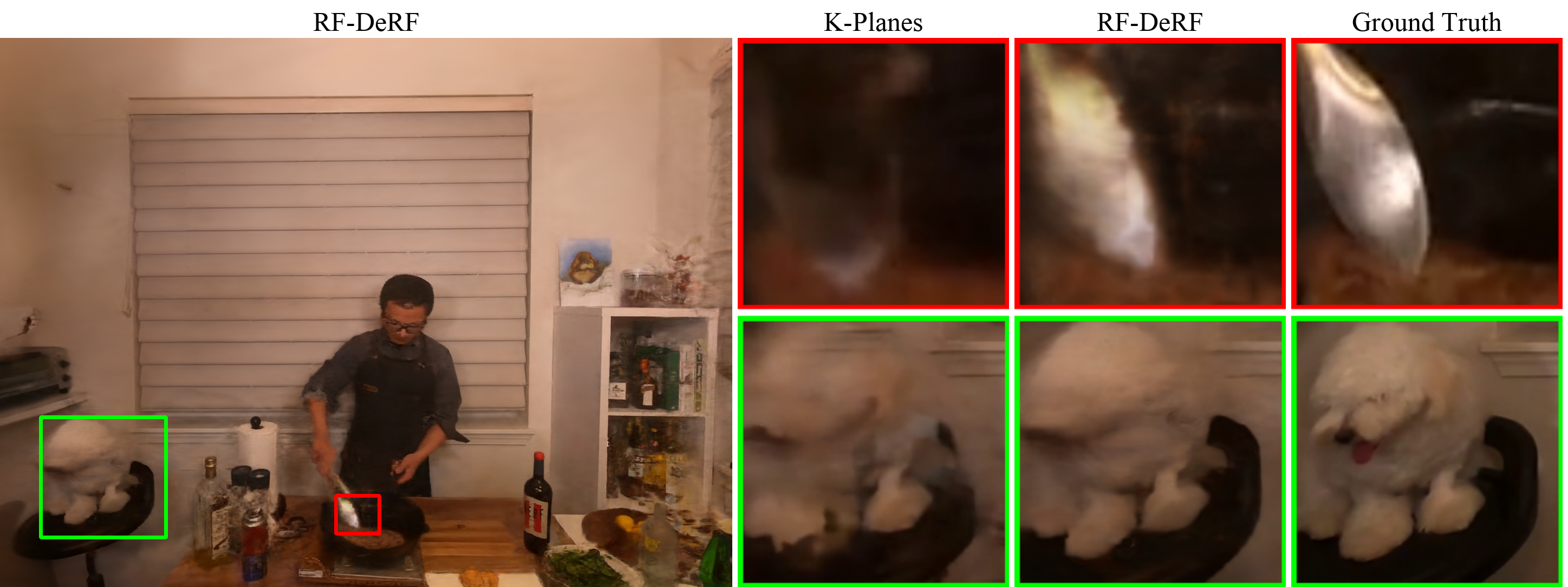}
    \caption{
        \textbf{Qualitative examples on N3DV dataset with four input views:}
        We find that K-Planes is unable to learn the moving tongs in the first row, while it struggles to learn the shape of the dog in the second row.
        Our model is able to reconstruct both, albeit with some blur.
    }
    \label{fig:qualitative-main-n3dv-4views}
\end{figure*}

%% file: tex/figures/qualitative_ablations_n3dv_3views.tex
\begin{figure*}
    \centering
    \begin{subfigure}[b]{0.48\linewidth}
        \centering
        \includegraphics[width=\linewidth]{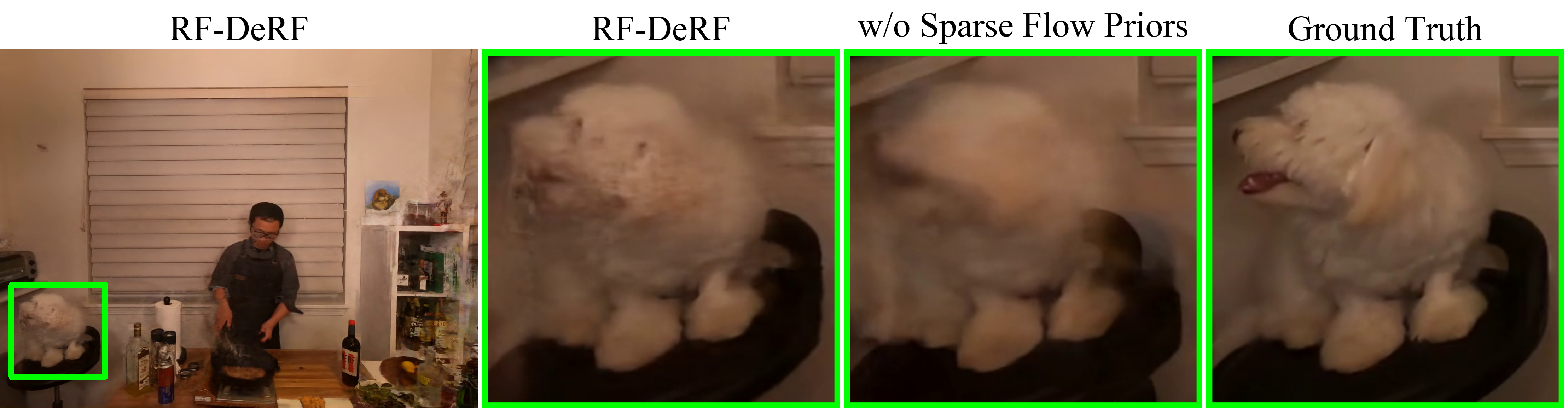}
        \caption{
            Without sparse flow priors, we observe that the face of the dog suffers from motion blur creating a fuzzy white mass.
        }
        \label{fig:qualitative-ablations-n3dv-3views-without-sf}
    \end{subfigure} \hfill
    \begin{subfigure}[b]{0.48\linewidth}
        \centering
        \includegraphics[width=\linewidth]{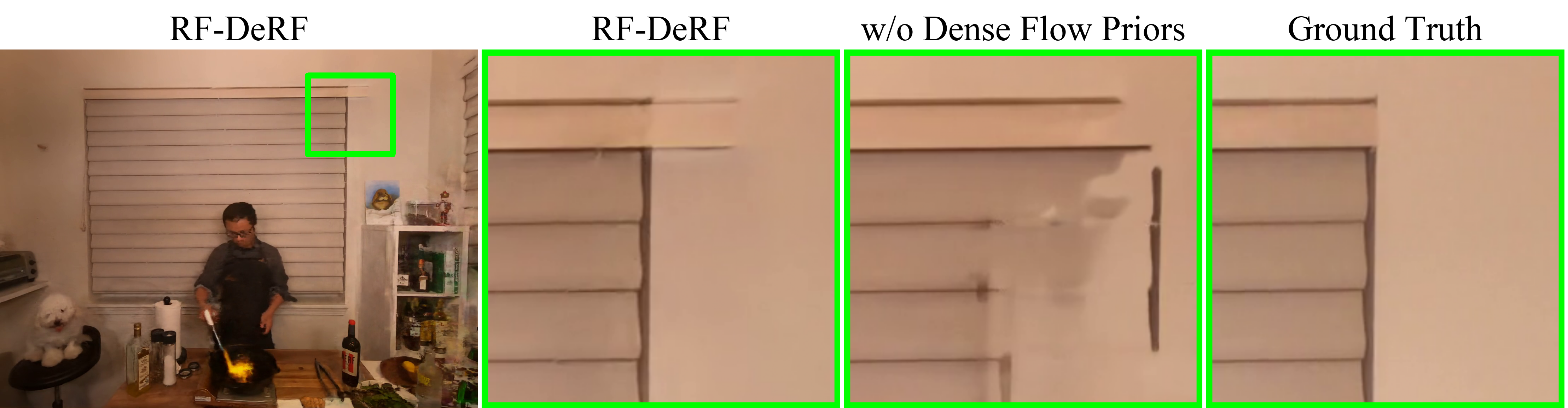}
        \caption{
            Absence of dense flow priors leads to distortions on the wall, where we observe the window curtain bleeding onto the wall.
        }
        \label{fig:qualitative-ablations-n3dv-3views-without-df}
    \end{subfigure}
    \caption{
        \textbf{Qualitative examples of ablated models on N3DV dataset:}
        Sparse flow prior is effective in regularizing moving regions in the scene.
        On the other hand, dense flow prior is effective in regularizing the smooth and static regions.
        Employing both the priors leads to the best reconstruction without either artifacts.
    }
    \label{fig:qualitative-ablations-n3dv-3views}
\end{figure*}

%% file: tex/figures/qualitative_main_id_2views.tex
\begin{figure*}
    \centering
    \includegraphics[width=\linewidth]{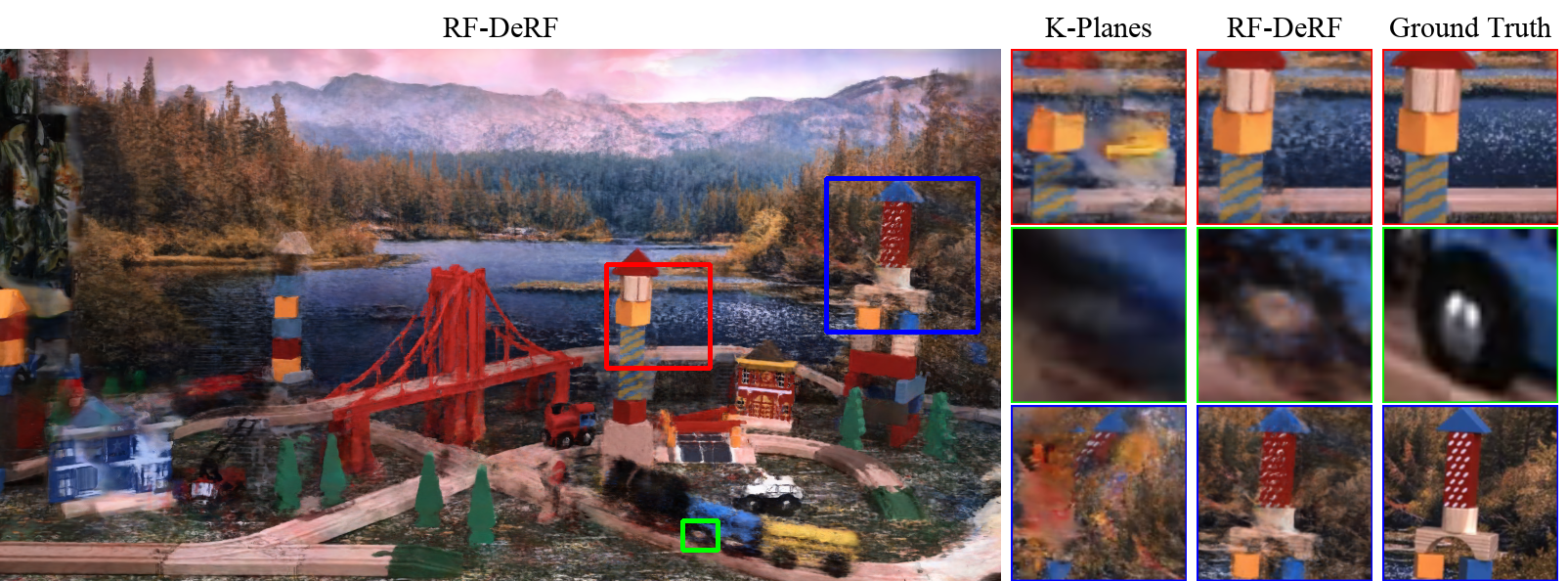}
    \caption{
        \textbf{Qualitative examples on InterDigital dataset:}
        In the first row, K-Planes creates a duplication of the tower, while the tower is significantly distorted in the third row.
        Our model correctly reconstructs both the towers and better reconstructs the wheel of the moving train.
    }
    \label{fig:qualitative-main-id-2views}
\end{figure*}

%% file: tex/figures/qualitative_depth_id_2views.tex
    \centering
    \includegraphics[width=0.95\linewidth]{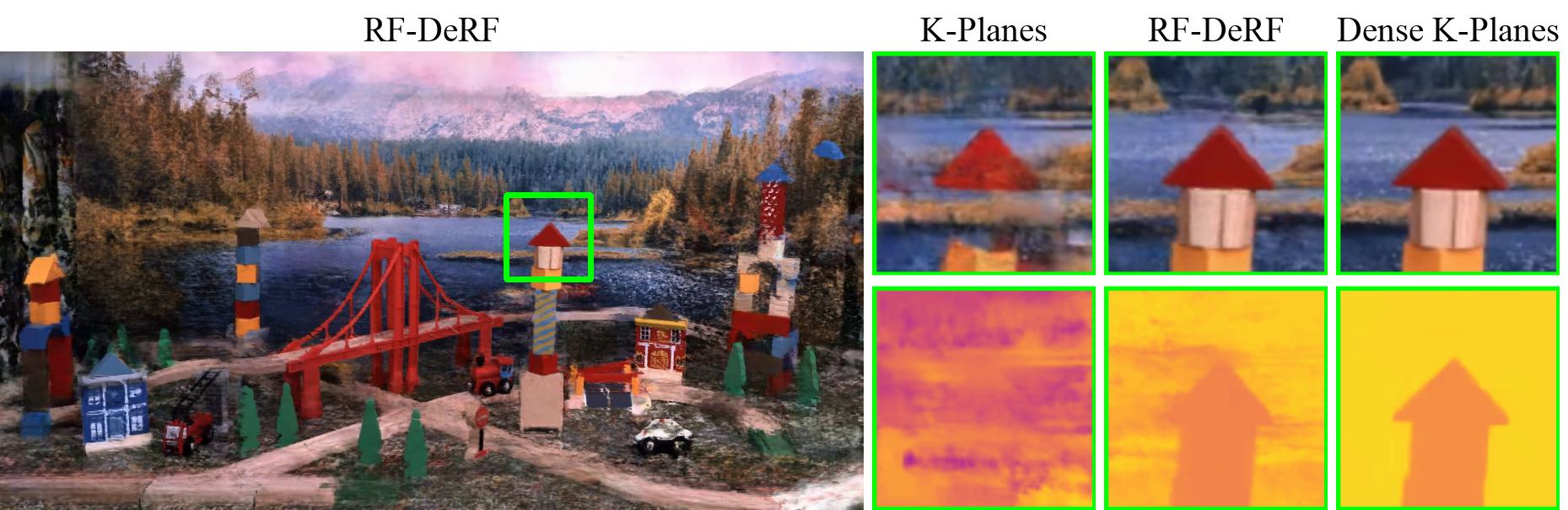}
    \caption{
        \textbf{Visualization of rendered depth on InterDigital dataset:}
        While our model learns better depth in the scene leading to better reconstruction of the tower, K-Planes is unable to learn the geometry correctly causing distortions in the tower.
    }
    \label{fig:qualitative-depth-id-2views}

%% file: tex/figures/qualitative_depth_n3dv_3views.tex
    \centering
    \includegraphics[width=\linewidth]{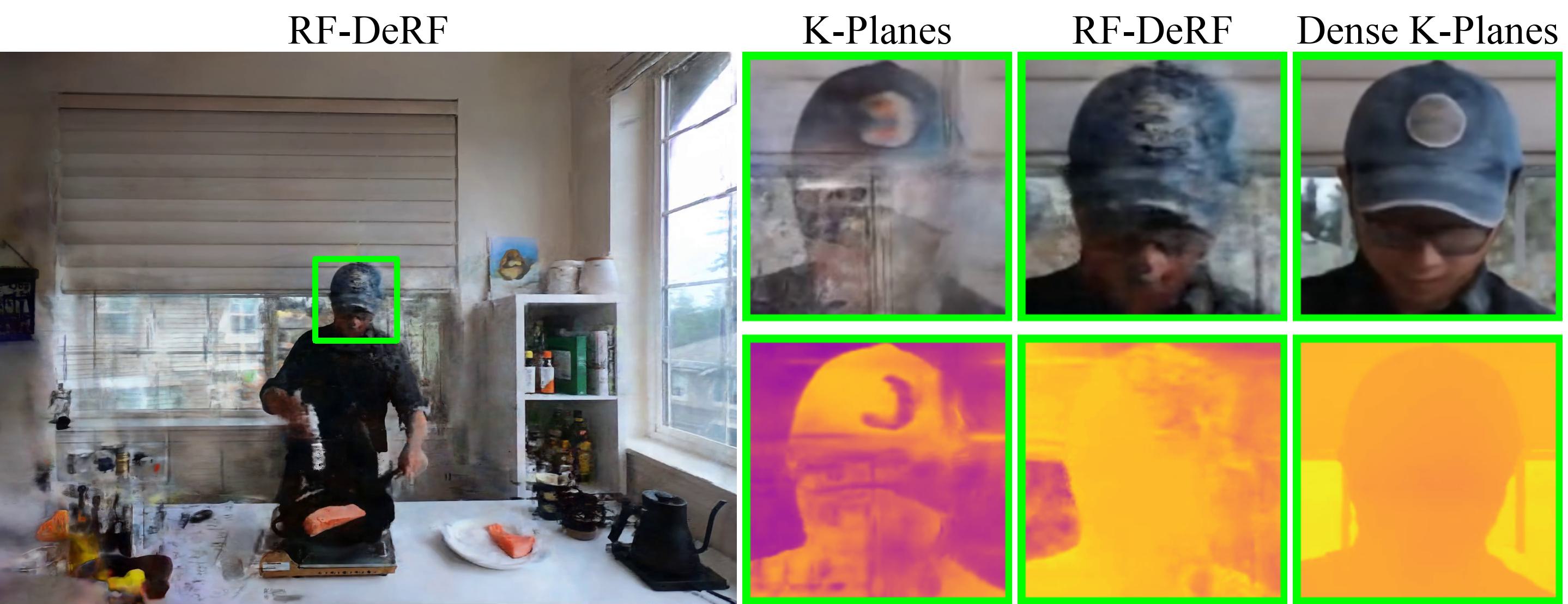}
    \caption{
        \textbf{Visualization of rendered depth on N3DV dataset:}
        Observe the difference in color of the depth map rendered by K-Planes which shows the errors in the estimated depth.
    }
    \label{fig:qualitative-depth-n3dv-3views}

%% file: tex/figures/qualitative_naive_flow_n3dv_3views.tex
    \centering
    \includegraphics[width=0.95\linewidth]{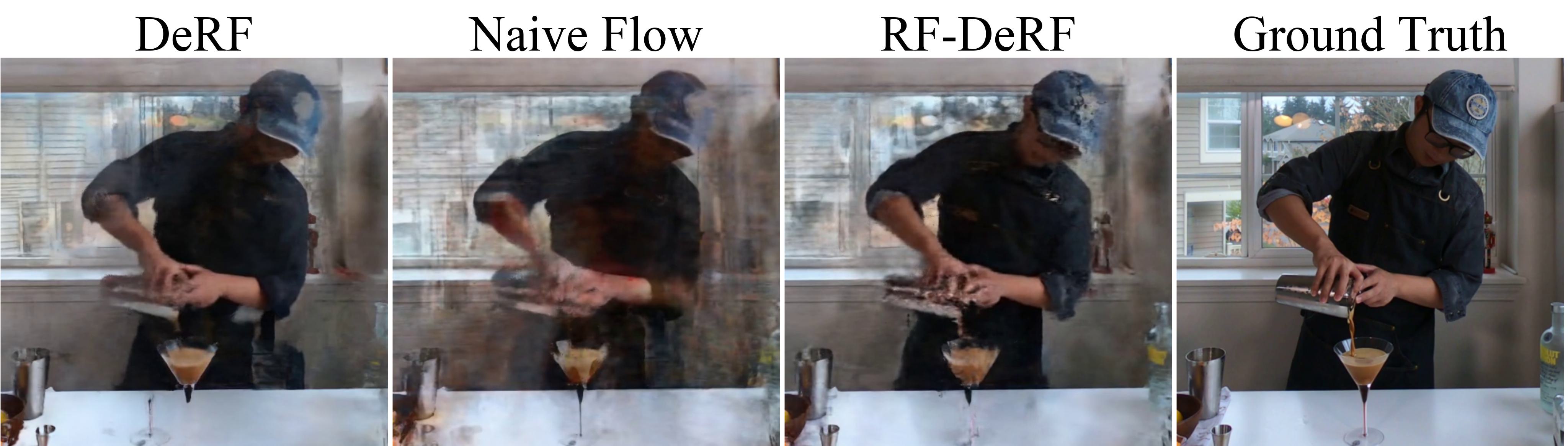}
    \caption{
        \textbf{Qualitative examples to show the effect of naive cross-camera dense flow priors:}
        We observe that the hand of the person in the second column is distorted, perhaps due to incorrect priors provided by the cross-camera dense flow.
        Further, we observe that the cross-camera dense flow priors deteriorate the performance as compared to our base DeRF model.
        Our priors do not cause such distortions, while improving the overall reconstruction quality (observe the color of the shirt).
    }
    \label{fig:qualitative-naive-flow-n3dv-3views}

%% file: tex/figures/quantitative_increasing_views_n3dv.tex
    \centering
    \includegraphics[width=0.9\linewidth]{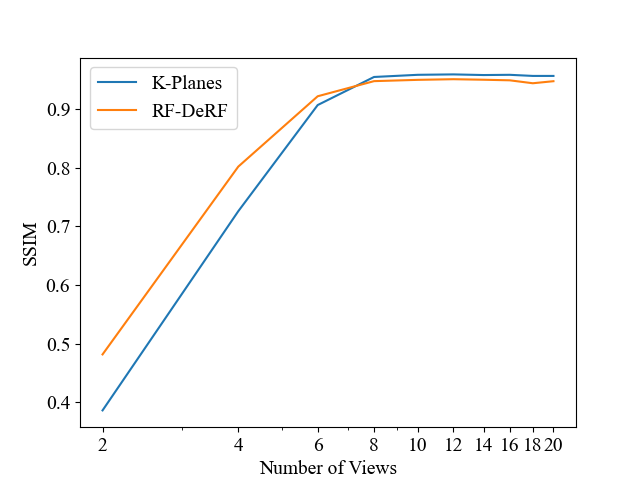}
    \caption{
        \textbf{Quantitative results on the flame salmon scene of N3DV dataset with different number of input views:}
        We observe that our model outperforms K-Planes in sparse input setting, while being competitive when more input views are available.
    }
    \label{fig:quantitative-increasing-views-n3dv}

%% file: DSSN.bbl

\begin{thebibliography}{81}


\ifx \showCODEN    \undefined \def \showCODEN     #1{\unskip}     \fi
\ifx \showDOI      \undefined \def \showDOI       #1{#1}\fi
\ifx \showISBNx    \undefined \def \showISBNx     #1{\unskip}     \fi
\ifx \showISBNxiii \undefined \def \showISBNxiii  #1{\unskip}     \fi
\ifx \showISSN     \undefined \def \showISSN      #1{\unskip}     \fi
\ifx \showLCCN     \undefined \def \showLCCN      #1{\unskip}     \fi
\ifx \shownote     \undefined \def \shownote      #1{#1}          \fi
\ifx \showarticletitle \undefined \def \showarticletitle #1{#1}   \fi
\ifx \showURL      \undefined \def \showURL       {\relax}        \fi
\providecommand\bibfield[2]{#2}
\providecommand\bibinfo[2]{#2}
\providecommand\natexlab[1]{#1}
\providecommand\showeprint[2][]{arXiv:#2}

\bibitem[Adrian(1991)]%
        {adrian1991particle}
\bibfield{author}{\bibinfo{person}{Ronald~J Adrian}.}
  \bibinfo{year}{1991}\natexlab{}.
\newblock \showarticletitle{Particle-Imaging Techniques for Experimental Fluid
  Mechanics}.
\newblock \bibinfo{journal}{\emph{Annual review of fluid mechanics}}
  \bibinfo{volume}{23}, \bibinfo{number}{1} (\bibinfo{year}{1991}),
  \bibinfo{pages}{261--304}.
\newblock


\bibitem[Allen et~al\mbox{.}(2003)]%
        {allen2003space}
\bibfield{author}{\bibinfo{person}{Brett Allen}, \bibinfo{person}{Brian
  Curless}, {and} \bibinfo{person}{Zoran Popovi\'{c}}.}
  \bibinfo{year}{2003}\natexlab{}.
\newblock \showarticletitle{The Space of Human Body Shapes: Reconstruction and
  Parameterization from Range Scans}.
\newblock \bibinfo{journal}{\emph{ACM Transactions on Graphics (TOG)}}
  \bibinfo{volume}{22}, \bibinfo{number}{3} (\bibinfo{date}{July}
  \bibinfo{year}{2003}), \bibinfo{pages}{587--594}.
\newblock
\urldef\tempurl%
\url{https://doi.org/10.1145/882262.882311}
\showDOI{\tempurl}


\bibitem[Anguelov et~al\mbox{.}(2005)]%
        {anguelov2005scape}
\bibfield{author}{\bibinfo{person}{Dragomir Anguelov}, \bibinfo{person}{Praveen
  Srinivasan}, \bibinfo{person}{Daphne Koller}, \bibinfo{person}{Sebastian
  Thrun}, \bibinfo{person}{Jim Rodgers}, {and} \bibinfo{person}{James Davis}.}
  \bibinfo{year}{2005}\natexlab{}.
\newblock \showarticletitle{{SCAPE}: Shape Completion and Animation of People}.
  In \bibinfo{booktitle}{\emph{Proceedings of the ACM Conference on Computer
  Graphics and Interactive Techniques (SIGGRAPH)}}.
\newblock
\urldef\tempurl%
\url{https://doi.org/10.1145/1186822.1073207}
\showDOI{\tempurl}


\bibitem[Atcheson et~al\mbox{.}(2008)]%
        {atcheson2008time}
\bibfield{author}{\bibinfo{person}{Bradley Atcheson}, \bibinfo{person}{Ivo
  Ihrke}, \bibinfo{person}{Wolfgang Heidrich}, \bibinfo{person}{Art Tevs},
  \bibinfo{person}{Derek Bradley}, \bibinfo{person}{Marcus Magnor}, {and}
  \bibinfo{person}{Hans-Peter Seidel}.} \bibinfo{year}{2008}\natexlab{}.
\newblock \showarticletitle{Time-resolved {3D} Capture of Non-stationary Gas
  Flows}.
\newblock \bibinfo{journal}{\emph{ACM Transactions on Graphics (TOG)}}
  \bibinfo{volume}{27}, \bibinfo{number}{5} (\bibinfo{date}{December}
  \bibinfo{year}{2008}).
\newblock
\urldef\tempurl%
\url{https://doi.org/10.1145/1409060.1409085}
\showDOI{\tempurl}


\bibitem[Bansal and Zollh\"ofer(2023)]%
        {bansal2023npc}
\bibfield{author}{\bibinfo{person}{Aayush Bansal} {and}
  \bibinfo{person}{Michael Zollh\"ofer}.} \bibinfo{year}{2023}\natexlab{}.
\newblock \showarticletitle{Neural Pixel Composition for 3D-4D View Synthesis
  From Multi-Views}. In \bibinfo{booktitle}{\emph{Proceedings of the IEEE/CVF
  Conference on Computer Vision and Pattern Recognition (CVPR)}}.
\newblock


\bibitem[Barron et~al\mbox{.}(2023)]%
        {barron2023zipnerf}
\bibfield{author}{\bibinfo{person}{Jonathan~T. Barron}, \bibinfo{person}{Ben
  Mildenhall}, \bibinfo{person}{Dor Verbin}, \bibinfo{person}{Pratul~P.
  Srinivasan}, {and} \bibinfo{person}{Peter Hedman}.}
  \bibinfo{year}{2023}\natexlab{}.
\newblock \showarticletitle{{Zip-NeRF}: Anti-Aliased Grid-Based Neural Radiance
  Fields}. In \bibinfo{booktitle}{\emph{Proceedings of the IEEE/CVF
  International Conference on Computer Vision (ICCV)}}.
\newblock


\bibitem[Blanz and Vetter(1999)]%
        {blanz1999morphable}
\bibfield{author}{\bibinfo{person}{Volker Blanz} {and} \bibinfo{person}{Thomas
  Vetter}.} \bibinfo{year}{1999}\natexlab{}.
\newblock \showarticletitle{A morphable model for the synthesis of 3D faces}.
  In \bibinfo{booktitle}{\emph{Proceedings of the ACM Conference on Computer
  Graphics and Interactive Techniques (SIGGRAPH)}}.
\newblock
\urldef\tempurl%
\url{https://doi.org/10.1145/311535.311556}
\showDOI{\tempurl}


\bibitem[Bradley et~al\mbox{.}(2010)]%
        {bradley2010high}
\bibfield{author}{\bibinfo{person}{Derek Bradley}, \bibinfo{person}{Wolfgang
  Heidrich}, \bibinfo{person}{Tiberiu Popa}, {and} \bibinfo{person}{Alla
  Sheffer}.} \bibinfo{year}{2010}\natexlab{}.
\newblock \showarticletitle{High Resolution Passive Facial Performance
  Capture}. In \bibinfo{booktitle}{\emph{Proceedings of the ACM Conference on
  Computer Graphics and Interactive Techniques (SIGGRAPH)}}.
\newblock
\urldef\tempurl%
\url{https://doi.org/10.1145/1833349.1778778}
\showDOI{\tempurl}


\bibitem[Bradley et~al\mbox{.}(2008)]%
        {bradley2008markerless}
\bibfield{author}{\bibinfo{person}{Derek Bradley}, \bibinfo{person}{Tiberiu
  Popa}, \bibinfo{person}{Alla Sheffer}, \bibinfo{person}{Wolfgang Heidrich},
  {and} \bibinfo{person}{Tamy Boubekeur}.} \bibinfo{year}{2008}\natexlab{}.
\newblock \showarticletitle{Markerless Garment Capture}. In
  \bibinfo{booktitle}{\emph{Proceedings of the ACM Conference on Computer
  Graphics and Interactive Techniques (SIGGRAPH)}}.
\newblock
\urldef\tempurl%
\url{https://doi.org/10.1145/1399504.1360698}
\showDOI{\tempurl}


\bibitem[Cao and Johnson(2023)]%
        {cao2023hexplane}
\bibfield{author}{\bibinfo{person}{Ang Cao} {and} \bibinfo{person}{Justin
  Johnson}.} \bibinfo{year}{2023}\natexlab{}.
\newblock \showarticletitle{{HexPlane}: A Fast Representation for Dynamic
  Scenes}. In \bibinfo{booktitle}{\emph{Proceedings of the IEEE/CVF Conference
  on Computer Vision and Pattern Recognition (CVPR)}}.
\newblock


\bibitem[Carranza et~al\mbox{.}(2003)]%
        {carranza2003free}
\bibfield{author}{\bibinfo{person}{Joel Carranza}, \bibinfo{person}{Christian
  Theobalt}, \bibinfo{person}{Marcus~A. Magnor}, {and}
  \bibinfo{person}{Hans-Peter Seidel}.} \bibinfo{year}{2003}\natexlab{}.
\newblock \showarticletitle{Free-Viewpoint Video of Human Actors}.
\newblock \bibinfo{journal}{\emph{ACM Transactions on Graphics (TOG)}}
  \bibinfo{volume}{22}, \bibinfo{number}{3} (\bibinfo{date}{July}
  \bibinfo{year}{2003}), \bibinfo{pages}{569--577}.
\newblock
\urldef\tempurl%
\url{https://doi.org/10.1145/882262.882309}
\showDOI{\tempurl}


\bibitem[Chen et~al\mbox{.}(2022)]%
        {chen2022tensorf}
\bibfield{author}{\bibinfo{person}{Anpei Chen}, \bibinfo{person}{Zexiang Xu},
  \bibinfo{person}{Andreas Geiger}, \bibinfo{person}{Jingyi Yu}, {and}
  \bibinfo{person}{Hao Su}.} \bibinfo{year}{2022}\natexlab{}.
\newblock \showarticletitle{{TensoRF}: Tensorial Radiance Fields}. In
  \bibinfo{booktitle}{\emph{Proceedings of the European Conference on Computer
  Vision (ECCV)}}.
\newblock


\bibitem[de~Aguiar et~al\mbox{.}(2008)]%
        {de2008performance}
\bibfield{author}{\bibinfo{person}{Edilson de Aguiar}, \bibinfo{person}{Carsten
  Stoll}, \bibinfo{person}{Christian Theobalt}, \bibinfo{person}{Naveed Ahmed},
  \bibinfo{person}{Hans-Peter Seidel}, {and} \bibinfo{person}{Sebastian
  Thrun}.} \bibinfo{year}{2008}\natexlab{}.
\newblock \showarticletitle{Performance Capture from Sparse Multi-View Video}.
  In \bibinfo{booktitle}{\emph{Proceedings of the ACM Conference on Computer
  Graphics and Interactive Techniques (SIGGRAPH)}}.
\newblock
\urldef\tempurl%
\url{https://doi.org/10.1145/1399504.1360697}
\showDOI{\tempurl}


\bibitem[Deng et~al\mbox{.}(2022)]%
        {deng2022dsnerf}
\bibfield{author}{\bibinfo{person}{Kangle Deng}, \bibinfo{person}{Andrew Liu},
  \bibinfo{person}{Jun-Yan Zhu}, {and} \bibinfo{person}{Deva Ramanan}.}
  \bibinfo{year}{2022}\natexlab{}.
\newblock \showarticletitle{Depth-Supervised {NeRF}: Fewer Views and Faster
  Training for Free}. In \bibinfo{booktitle}{\emph{Proceedings of the IEEE/CVF
  Conference on Computer Vision and Pattern Recognition (CVPR)}}.
\newblock


\bibitem[Faloutsos et~al\mbox{.}(1997)]%
        {faloutsos1997dynamic}
\bibfield{author}{\bibinfo{person}{P. Faloutsos}, \bibinfo{person}{M. Van~de
  Panne}, {and} \bibinfo{person}{D. Terzopoulos}.}
  \bibinfo{year}{1997}\natexlab{}.
\newblock \showarticletitle{Dynamic Free-Form Deformations for Animation
  Synthesis}.
\newblock \bibinfo{journal}{\emph{IEEE Transactions on Visualization and
  Computer Graphics (TVCG)}} \bibinfo{volume}{3}, \bibinfo{number}{3}
  (\bibinfo{year}{1997}), \bibinfo{pages}{201--214}.
\newblock
\urldef\tempurl%
\url{https://doi.org/10.1109/2945.620488}
\showDOI{\tempurl}


\bibitem[Fang et~al\mbox{.}(2022)]%
        {fang2022tineuvox}
\bibfield{author}{\bibinfo{person}{Jiemin Fang}, \bibinfo{person}{Taoran Yi},
  \bibinfo{person}{Xinggang Wang}, \bibinfo{person}{Lingxi Xie},
  \bibinfo{person}{Xiaopeng Zhang}, \bibinfo{person}{Wenyu Liu},
  \bibinfo{person}{Matthias Nie\ss{}ner}, {and} \bibinfo{person}{Qi Tian}.}
  \bibinfo{year}{2022}\natexlab{}.
\newblock \showarticletitle{Fast Dynamic Radiance Fields with Time-Aware Neural
  Voxels}. In \bibinfo{booktitle}{\emph{Proceedings of the SIGGRAPH Asia 2022
  Conference Papers}}.
\newblock
\urldef\tempurl%
\url{https://doi.org/10.1145/3550469.3555383}
\showDOI{\tempurl}


\bibitem[Fridovich-Keil et~al\mbox{.}(2023)]%
        {fridovich2023kplanes}
\bibfield{author}{\bibinfo{person}{Sara Fridovich-Keil},
  \bibinfo{person}{Giacomo Meanti}, \bibinfo{person}{Frederik~Rahb{\ae}k
  Warburg}, \bibinfo{person}{Benjamin Recht}, {and} \bibinfo{person}{Angjoo
  Kanazawa}.} \bibinfo{year}{2023}\natexlab{}.
\newblock \showarticletitle{{K-Planes}: Explicit Radiance Fields in Space,
  Time, and Appearance}. In \bibinfo{booktitle}{\emph{Proceedings of the
  IEEE/CVF Conference on Computer Vision and Pattern Recognition (CVPR)}}.
\newblock


\bibitem[Fridovich-Keil et~al\mbox{.}(2022)]%
        {fridovich2022plenoxels}
\bibfield{author}{\bibinfo{person}{Sara Fridovich-Keil}, \bibinfo{person}{Alex
  Yu}, \bibinfo{person}{Matthew Tancik}, \bibinfo{person}{Qinhong Chen},
  \bibinfo{person}{Benjamin Recht}, {and} \bibinfo{person}{Angjoo Kanazawa}.}
  \bibinfo{year}{2022}\natexlab{}.
\newblock \showarticletitle{{Plenoxels}: Radiance Fields Without Neural
  Networks}. In \bibinfo{booktitle}{\emph{Proceedings of the IEEE/CVF
  Conference on Computer Vision and Pattern Recognition (CVPR)}}.
\newblock


\bibitem[Gregson et~al\mbox{.}(2012)]%
        {gregson2012stochastic}
\bibfield{author}{\bibinfo{person}{James Gregson}, \bibinfo{person}{Michael
  Krimerman}, \bibinfo{person}{Matthias~B Hullin}, {and}
  \bibinfo{person}{Wolfgang Heidrich}.} \bibinfo{year}{2012}\natexlab{}.
\newblock \showarticletitle{Stochastic Tomography and its Applications in 3D
  Imaging of Mixing Fluids}.
\newblock \bibinfo{journal}{\emph{ACM Transactions on Graphics (TOG)}}
  \bibinfo{volume}{31}, \bibinfo{number}{4} (\bibinfo{date}{July}
  \bibinfo{year}{2012}), \bibinfo{pages}{1--10}.
\newblock


\bibitem[Guo et~al\mbox{.}(2023)]%
        {guo2023forward}
\bibfield{author}{\bibinfo{person}{Xiang Guo}, \bibinfo{person}{Jiadai Sun},
  \bibinfo{person}{Yuchao Dai}, \bibinfo{person}{Guanying Chen},
  \bibinfo{person}{Xiaoqing Ye}, \bibinfo{person}{Xiao Tan},
  \bibinfo{person}{Errui Ding}, \bibinfo{person}{Yumeng Zhang}, {and}
  \bibinfo{person}{Jingdong Wang}.} \bibinfo{year}{2023}\natexlab{}.
\newblock \showarticletitle{Forward Flow for Novel View Synthesis of Dynamic
  Scenes}. In \bibinfo{booktitle}{\emph{Proceedings of the IEEE/CVF
  International Conference on Computer Vision (ICCV)}}.
\newblock


\bibitem[Guskov et~al\mbox{.}(2003)]%
        {guskov2003trackable}
\bibfield{author}{\bibinfo{person}{Igor Guskov}, \bibinfo{person}{Sergey
  Klibanov}, {and} \bibinfo{person}{Benjamin Bryant}.}
  \bibinfo{year}{2003}\natexlab{}.
\newblock \showarticletitle{Trackable Surfaces}. In
  \bibinfo{booktitle}{\emph{Proceedings of the ACM SIGGRAPH/Eurographics
  Symposium on Computer Animation (SCA)}}.
\newblock


\bibitem[Hasler et~al\mbox{.}(2006)]%
        {hasler2006physically}
\bibfield{author}{\bibinfo{person}{Nils Hasler}, \bibinfo{person}{Mark Asbach},
  \bibinfo{person}{Bodo Rosenhahn}, \bibinfo{person}{Jens-Rainer Ohm}, {and}
  \bibinfo{person}{Hans-Peter Seidel}.} \bibinfo{year}{2006}\natexlab{}.
\newblock \showarticletitle{Physically based Tracking of Cloth}. In
  \bibinfo{booktitle}{\emph{Proceedings of the International Workshop on
  Vision, Modeling, and Visualization (VMV)}}.
\newblock


\bibitem[Hawkins et~al\mbox{.}(2005)]%
        {hawkins2005acquisition}
\bibfield{author}{\bibinfo{person}{Tim Hawkins}, \bibinfo{person}{Per
  Einarsson}, {and} \bibinfo{person}{Paul Debevec}.}
  \bibinfo{year}{2005}\natexlab{}.
\newblock \showarticletitle{Acquisition of Time-Varying Participating Media}.
\newblock \bibinfo{journal}{\emph{ACM Transactions on Graphics (TOG)}}
  \bibinfo{volume}{24}, \bibinfo{number}{3} (\bibinfo{date}{July}
  \bibinfo{year}{2005}), \bibinfo{pages}{812--815}.
\newblock
\urldef\tempurl%
\url{https://doi.org/10.1145/1073204.1073266}
\showDOI{\tempurl}


\bibitem[Ihrke and Magnor(2004)]%
        {ihrke2004image}
\bibfield{author}{\bibinfo{person}{Ivo Ihrke} {and} \bibinfo{person}{Marcus
  Magnor}.} \bibinfo{year}{2004}\natexlab{}.
\newblock \showarticletitle{Image-Based Tomographic Reconstruction of Flames}.
  In \bibinfo{booktitle}{\emph{Proceedings of the ACM SIGGRAPH/Eurographics
  Symposium on Computer Animation (SCA)}}.
\newblock
\urldef\tempurl%
\url{https://doi.org/10.1145/1028523.1028572}
\showDOI{\tempurl}


\bibitem[Jain et~al\mbox{.}(2021)]%
        {jain2021dietnerf}
\bibfield{author}{\bibinfo{person}{Ajay Jain}, \bibinfo{person}{Matthew
  Tancik}, {and} \bibinfo{person}{Pieter Abbeel}.}
  \bibinfo{year}{2021}\natexlab{}.
\newblock \showarticletitle{Putting {NeRF} on a Diet: Semantically Consistent
  Few-Shot View Synthesis}. In \bibinfo{booktitle}{\emph{Proceedings of the
  IEEE/CVF International Conference on Computer Vision (ICCV)}}.
\newblock


\bibitem[Joo et~al\mbox{.}(2018)]%
        {joo2018total}
\bibfield{author}{\bibinfo{person}{Hanbyul Joo}, \bibinfo{person}{Tomas Simon},
  {and} \bibinfo{person}{Yaser Sheikh}.} \bibinfo{year}{2018}\natexlab{}.
\newblock \showarticletitle{Total Capture: A 3D Deformation Model for Tracking
  Faces, Hands, and Bodies}. In \bibinfo{booktitle}{\emph{Proceedings of the
  IEEE Conference on Computer Vision and Pattern Recognition (CVPR)}}.
\newblock


\bibitem[Kerbl et~al\mbox{.}(2023)]%
        {kerbl20233dgs}
\bibfield{author}{\bibinfo{person}{Bernhard Kerbl}, \bibinfo{person}{Georgios
  Kopanas}, \bibinfo{person}{Thomas Leimk{\"u}hler}, {and}
  \bibinfo{person}{George Drettakis}.} \bibinfo{year}{2023}\natexlab{}.
\newblock \showarticletitle{3D Gaussian Splatting for Real-Time Radiance Field
  Rendering}.
\newblock \bibinfo{journal}{\emph{ACM Transactions on Graphics (TOG)}}
  \bibinfo{volume}{42}, \bibinfo{number}{4} (\bibinfo{year}{2023}).
\newblock


\bibitem[Kim et~al\mbox{.}(2022)]%
        {kim2022infonerf}
\bibfield{author}{\bibinfo{person}{Mijeong Kim}, \bibinfo{person}{Seonguk Seo},
  {and} \bibinfo{person}{Bohyung Han}.} \bibinfo{year}{2022}\natexlab{}.
\newblock \showarticletitle{{InfoNeRF}: Ray Entropy Minimization for Few-Shot
  Neural Volume Rendering}. In \bibinfo{booktitle}{\emph{Proceedings of the
  IEEE/CVF Conference on Computer Vision and Pattern Recognition (CVPR)}}.
\newblock


\bibitem[Kraevoy and Sheffer(2004)]%
        {kraevoy2004cross}
\bibfield{author}{\bibinfo{person}{Vladislav Kraevoy} {and}
  \bibinfo{person}{Alla Sheffer}.} \bibinfo{year}{2004}\natexlab{}.
\newblock \showarticletitle{Cross-Parameterization and Compatible Remeshing of
  3D Models}.
\newblock \bibinfo{journal}{\emph{ACM Transactions on Graphics (TOG)}}
  \bibinfo{volume}{23}, \bibinfo{number}{3} (\bibinfo{date}{August}
  \bibinfo{year}{2004}), \bibinfo{pages}{861--869}.
\newblock
\urldef\tempurl%
\url{https://doi.org/10.1145/1015706.1015811}
\showDOI{\tempurl}


\bibitem[Kraevoy and Sheffer(2005)]%
        {kraevoy2005template}
\bibfield{author}{\bibinfo{person}{Vladislav Kraevoy} {and}
  \bibinfo{person}{Alla Sheffer}.} \bibinfo{year}{2005}\natexlab{}.
\newblock \showarticletitle{Template-Based Mesh Completion}. In
  \bibinfo{booktitle}{\emph{Proceedings of the Symposium on Geometry Processing
  (SGP)}}.
\newblock


\bibitem[Lee et~al\mbox{.}(2023)]%
        {lee2023fast}
\bibfield{author}{\bibinfo{person}{Yao-Chih Lee}, \bibinfo{person}{Zhoutong
  Zhang}, \bibinfo{person}{Kevin Blackburn-Matzen}, \bibinfo{person}{Simon
  Niklaus}, \bibinfo{person}{Jianming Zhang}, \bibinfo{person}{Jia-Bin Huang},
  {and} \bibinfo{person}{Feng Liu}.} \bibinfo{year}{2023}\natexlab{}.
\newblock \showarticletitle{Fast View Synthesis of Casual Videos}.
\newblock \bibinfo{journal}{\emph{arXiv e-prints}}, Article
  \bibinfo{articleno}{arXiv:2312.02135} (\bibinfo{year}{2023}),
  \bibinfo{numpages}{arXiv:2312.02135}~pages.
\newblock
\showeprint[arxiv]{2312.02135}


\bibitem[Li et~al\mbox{.}(2023)]%
        {li2023dynamic}
\bibfield{author}{\bibinfo{person}{Deqi Li}, \bibinfo{person}{Shi-Sheng Huang},
  \bibinfo{person}{Tianyu Shen}, {and} \bibinfo{person}{Hua Huang}.}
  \bibinfo{year}{2023}\natexlab{}.
\newblock \showarticletitle{Dynamic View Synthesis with Spatio-Temporal Feature
  Warping from Sparse Views}. In \bibinfo{booktitle}{\emph{Proceedings of the
  ACM International Conference on Multimedia (ACM-MM)}}.
\newblock
\urldef\tempurl%
\url{https://doi.org/10.1145/3581783.3612419}
\showDOI{\tempurl}


\bibitem[Li et~al\mbox{.}(2022)]%
        {li2022neural}
\bibfield{author}{\bibinfo{person}{Tianye Li}, \bibinfo{person}{Mira
  Slavcheva}, \bibinfo{person}{Michael Zollh\"ofer}, \bibinfo{person}{Simon
  Green}, \bibinfo{person}{Christoph Lassner}, \bibinfo{person}{Changil Kim},
  \bibinfo{person}{Tanner Schmidt}, \bibinfo{person}{Steven Lovegrove},
  \bibinfo{person}{Michael Goesele}, \bibinfo{person}{Richard Newcombe}, {and}
  \bibinfo{person}{Zhaoyang Lv}.} \bibinfo{year}{2022}\natexlab{}.
\newblock \showarticletitle{Neural {3D} Video Synthesis From Multi-View Video}.
  In \bibinfo{booktitle}{\emph{Proceedings of the IEEE/CVF Conference on
  Computer Vision and Pattern Recognition (CVPR)}}.
\newblock


\bibitem[Li et~al\mbox{.}(2021)]%
        {li2021nsff}
\bibfield{author}{\bibinfo{person}{Zhengqi Li}, \bibinfo{person}{Simon
  Niklaus}, \bibinfo{person}{Noah Snavely}, {and} \bibinfo{person}{Oliver
  Wang}.} \bibinfo{year}{2021}\natexlab{}.
\newblock \showarticletitle{Neural Scene Flow Fields for Space-Time View
  Synthesis of Dynamic Scenes}. In \bibinfo{booktitle}{\emph{Proceedings of the
  IEEE Conference on Computer Vision and Pattern Recognition (CVPR)}}.
\newblock


\bibitem[Lin et~al\mbox{.}(2021)]%
        {lin2021deep}
\bibfield{author}{\bibinfo{person}{Kai-En Lin}, \bibinfo{person}{Lei Xiao},
  \bibinfo{person}{Feng Liu}, \bibinfo{person}{Guowei Yang}, {and}
  \bibinfo{person}{Ravi Ramamoorthi}.} \bibinfo{year}{2021}\natexlab{}.
\newblock \showarticletitle{Deep 3D Mask Volume for View Synthesis of Dynamic
  Scenes}. In \bibinfo{booktitle}{\emph{Proceedings of the IEEE International
  Conference on Computer Vision (ICCV)}}.
\newblock


\bibitem[Liu et~al\mbox{.}(2022)]%
        {liu2022devrf}
\bibfield{author}{\bibinfo{person}{Jia-Wei Liu}, \bibinfo{person}{Yan-Pei Cao},
  \bibinfo{person}{Weijia Mao}, \bibinfo{person}{Wenqiao Zhang},
  \bibinfo{person}{David~Junhao Zhang}, \bibinfo{person}{Jussi Keppo},
  \bibinfo{person}{Ying Shan}, \bibinfo{person}{Xiaohu Qie}, {and}
  \bibinfo{person}{Mike~Zheng Shou}.} \bibinfo{year}{2022}\natexlab{}.
\newblock \showarticletitle{DeVRF: Fast Deformable Voxel Radiance Fields for
  Dynamic Scenes}. In \bibinfo{booktitle}{\emph{Proceedings of the Advances in
  Neural Information Processing Systems (NeurIPS)}}.
\newblock


\bibitem[Liu et~al\mbox{.}(2020)]%
        {liu2020arflow}
\bibfield{author}{\bibinfo{person}{Liang Liu}, \bibinfo{person}{Jiangning
  Zhang}, \bibinfo{person}{Ruifei He}, \bibinfo{person}{Yong Liu},
  \bibinfo{person}{Yabiao Wang}, \bibinfo{person}{Ying Tai},
  \bibinfo{person}{Donghao Luo}, \bibinfo{person}{Chengjie Wang},
  \bibinfo{person}{Jilin Li}, {and} \bibinfo{person}{Feiyue Huang}.}
  \bibinfo{year}{2020}\natexlab{}.
\newblock \showarticletitle{Learning by Analogy: Reliable Supervision From
  Transformations for Unsupervised Optical Flow Estimation}. In
  \bibinfo{booktitle}{\emph{Proceedings of the IEEE Conference on Computer
  Vision and Pattern Recognition (CVPR)}}.
\newblock


\bibitem[Loper et~al\mbox{.}(2015)]%
        {loper2015smpl}
\bibfield{author}{\bibinfo{person}{Matthew Loper}, \bibinfo{person}{Naureen
  Mahmood}, \bibinfo{person}{Javier Romero}, \bibinfo{person}{Gerard
  Pons-Moll}, {and} \bibinfo{person}{Michael~J. Black}.}
  \bibinfo{year}{2015}\natexlab{}.
\newblock \showarticletitle{{SMPL}: A Skinned Multi-Person Linear Model}.
\newblock \bibinfo{journal}{\emph{ACM Transactions on Graphics (TOG)}}
  \bibinfo{volume}{34}, \bibinfo{number}{6} (\bibinfo{date}{October}
  \bibinfo{year}{2015}).
\newblock
\urldef\tempurl%
\url{https://doi.org/10.1145/2816795.2818013}
\showDOI{\tempurl}


\bibitem[Lowe(2004)]%
        {lowe2004sift}
\bibfield{author}{\bibinfo{person}{David~G Lowe}.}
  \bibinfo{year}{2004}\natexlab{}.
\newblock \showarticletitle{Distinctive Image Features from Scale-Invariant
  Keypoints}.
\newblock \bibinfo{journal}{\emph{International Journal of Computer Vision
  (IJCV)}}  \bibinfo{volume}{60} (\bibinfo{year}{2004}),
  \bibinfo{pages}{91--110}.
\newblock


\bibitem[Miguel et~al\mbox{.}(2012)]%
        {miguel2012data}
\bibfield{author}{\bibinfo{person}{Eder Miguel}, \bibinfo{person}{Derek
  Bradley}, \bibinfo{person}{Bernhard Thomaszewski}, \bibinfo{person}{Bernd
  Bickel}, \bibinfo{person}{Wojciech Matusik}, \bibinfo{person}{Miguel~A
  Otaduy}, {and} \bibinfo{person}{Steve Marschner}.}
  \bibinfo{year}{2012}\natexlab{}.
\newblock \showarticletitle{Data-Driven Estimation of Cloth Simulation Models}.
\newblock \bibinfo{journal}{\emph{Computer Graphics Forum (CGF)}}
  \bibinfo{volume}{31}, \bibinfo{number}{2pt2} (\bibinfo{year}{2012}),
  \bibinfo{pages}{519--528}.
\newblock
\urldef\tempurl%
\url{https://doi.org/10.1111/j.1467-8659.2012.03031.x}
\showDOI{\tempurl}


\bibitem[Mildenhall et~al\mbox{.}(2020)]%
        {mildenhall2020nerf}
\bibfield{author}{\bibinfo{person}{Ben Mildenhall}, \bibinfo{person}{Pratul~P.
  Srinivasan}, \bibinfo{person}{Matthew Tancik}, \bibinfo{person}{Jonathan~T.
  Barron}, \bibinfo{person}{Ravi Ramamoorthi}, {and} \bibinfo{person}{Ren Ng}.}
  \bibinfo{year}{2020}\natexlab{}.
\newblock \showarticletitle{NeRF: Representing Scenes as Neural Radiance Fields
  for View Synthesis}. In \bibinfo{booktitle}{\emph{Proceedings of the European
  Conference on Computer Vision (ECCV)}}.
\newblock


\bibitem[Mirzaei et~al\mbox{.}(2023)]%
        {mirzaei2023spinnerf}
\bibfield{author}{\bibinfo{person}{Ashkan Mirzaei}, \bibinfo{person}{Tristan
  Aumentado-Armstrong}, \bibinfo{person}{Konstantinos~G. Derpanis},
  \bibinfo{person}{Jonathan Kelly}, \bibinfo{person}{Marcus~A. Brubaker},
  \bibinfo{person}{Igor Gilitschenski}, {and} \bibinfo{person}{Alex
  Levinshtein}.} \bibinfo{year}{2023}\natexlab{}.
\newblock \showarticletitle{SPIn-NeRF: Multiview Segmentation and Perceptual
  Inpainting With Neural Radiance Fields}. In
  \bibinfo{booktitle}{\emph{Proceedings of the IEEE/CVF Conference on Computer
  Vision and Pattern Recognition (CVPR)}}.
\newblock


\bibitem[M{\"u}ller et~al\mbox{.}(2022)]%
        {muller2022instant}
\bibfield{author}{\bibinfo{person}{Thomas M{\"u}ller}, \bibinfo{person}{Alex
  Evans}, \bibinfo{person}{Christoph Schied}, {and} \bibinfo{person}{Alexander
  Keller}.} \bibinfo{year}{2022}\natexlab{}.
\newblock \showarticletitle{Instant Neural Graphics Primitives with a
  Multiresolution Hash Encoding}.
\newblock \bibinfo{journal}{\emph{ACM Transactions on Graphics (TOG)}}
  \bibinfo{volume}{41}, \bibinfo{number}{4} (\bibinfo{year}{2022}),
  \bibinfo{pages}{1--15}.
\newblock


\bibitem[Niemeyer et~al\mbox{.}(2022)]%
        {niemeyer2022regnerf}
\bibfield{author}{\bibinfo{person}{Michael Niemeyer},
  \bibinfo{person}{Jonathan~T. Barron}, \bibinfo{person}{Ben Mildenhall},
  \bibinfo{person}{Mehdi S.~M. Sajjadi}, \bibinfo{person}{Andreas Geiger},
  {and} \bibinfo{person}{Noha Radwan}.} \bibinfo{year}{2022}\natexlab{}.
\newblock \showarticletitle{{RegNeRF}: Regularizing Neural Radiance Fields for
  View Synthesis From Sparse Inputs}. In \bibinfo{booktitle}{\emph{Proceedings
  of the IEEE/CVF Conference on Computer Vision and Pattern Recognition
  (CVPR)}}.
\newblock


\bibitem[Nunes et~al\mbox{.}(2012)]%
        {nunes2012imaging}
\bibfield{author}{\bibinfo{person}{Hilario Nunes}, \bibinfo{person}{Yurdagul
  Uzunhan}, \bibinfo{person}{Thomas Gille}, \bibinfo{person}{Christine
  Lamberto}, \bibinfo{person}{Dominique Valeyre}, {and}
  \bibinfo{person}{Pierre-Yves Brillet}.} \bibinfo{year}{2012}\natexlab{}.
\newblock \showarticletitle{Imaging of Sarcoidosis of the Airways and Lung
  Parenchyma and Correlation with Lung Function}.
\newblock \bibinfo{journal}{\emph{European Respiratory Journal}}
  \bibinfo{volume}{40}, \bibinfo{number}{3} (\bibinfo{year}{2012}),
  \bibinfo{pages}{750--765}.
\newblock
\urldef\tempurl%
\url{https://doi.org/10.1183/09031936.00025212}
\showDOI{\tempurl}


\bibitem[Penner and Zhang(2017)]%
        {penner2017soft3d}
\bibfield{author}{\bibinfo{person}{Eric Penner} {and} \bibinfo{person}{Li
  Zhang}.} \bibinfo{year}{2017}\natexlab{}.
\newblock \showarticletitle{Soft {3D} Reconstruction for View Synthesis}.
\newblock \bibinfo{journal}{\emph{ACM Transactions on Graphics (TOG)}}
  \bibinfo{volume}{36}, \bibinfo{number}{6} (\bibinfo{date}{November}
  \bibinfo{year}{2017}), \bibinfo{pages}{1--11}.
\newblock
\urldef\tempurl%
\url{https://doi.org/10.1145/3130800.3130855}
\showDOI{\tempurl}


\bibitem[Pons-Moll et~al\mbox{.}(2017)]%
        {pons2017clothcap}
\bibfield{author}{\bibinfo{person}{Gerard Pons-Moll}, \bibinfo{person}{Sergi
  Pujades}, \bibinfo{person}{Sonny Hu}, {and} \bibinfo{person}{Michael~J.
  Black}.} \bibinfo{year}{2017}\natexlab{}.
\newblock \showarticletitle{{ClothCap}: Seamless 4D Clothing Capture and
  Retargeting}.
\newblock \bibinfo{journal}{\emph{ACM Transactions on Graphics (TOG)}}
  \bibinfo{volume}{36}, \bibinfo{number}{4} (\bibinfo{date}{July}
  \bibinfo{year}{2017}).
\newblock
\urldef\tempurl%
\url{https://doi.org/10.1145/3072959.3073711}
\showDOI{\tempurl}


\bibitem[Pumarola et~al\mbox{.}(2021)]%
        {pumarola2021dnerf}
\bibfield{author}{\bibinfo{person}{Albert Pumarola}, \bibinfo{person}{Enric
  Corona}, \bibinfo{person}{Gerard Pons-Moll}, {and} \bibinfo{person}{Francesc
  Moreno-Noguer}.} \bibinfo{year}{2021}\natexlab{}.
\newblock \showarticletitle{{D-NeRF}: Neural Radiance Fields for Dynamic
  Scenes}. In \bibinfo{booktitle}{\emph{Proceedings of the IEEE/CVF Conference
  on Computer Vision and Pattern Recognition (CVPR)}}.
\newblock


\bibitem[Revaud et~al\mbox{.}(2019)]%
        {revaud2019r2d2}
\bibfield{author}{\bibinfo{person}{Jerome Revaud}, \bibinfo{person}{Cesar
  De~Souza}, \bibinfo{person}{Martin Humenberger}, {and}
  \bibinfo{person}{Philippe Weinzaepfel}.} \bibinfo{year}{2019}\natexlab{}.
\newblock \showarticletitle{{R2D2}: Reliable and Repeatable Detector and
  Descriptor}. In \bibinfo{booktitle}{\emph{Proceedings of the Advances in
  Neural Information Processing Systems (NeurIPS)}}.
\newblock


\bibitem[Roessle et~al\mbox{.}(2022)]%
        {roessle2022ddpnerf}
\bibfield{author}{\bibinfo{person}{Barbara Roessle},
  \bibinfo{person}{Jonathan~T. Barron}, \bibinfo{person}{Ben Mildenhall},
  \bibinfo{person}{Pratul~P. Srinivasan}, {and} \bibinfo{person}{Matthias
  Nie{\ss}ner}.} \bibinfo{year}{2022}\natexlab{}.
\newblock \showarticletitle{Dense Depth Priors for Neural Radiance Fields From
  Sparse Input Views}. In \bibinfo{booktitle}{\emph{Proceedings of the IEEE/CVF
  Conference on Computer Vision and Pattern Recognition (CVPR)}}.
\newblock


\bibitem[Sabater et~al\mbox{.}(2017)]%
        {sabater2017interdigital}
\bibfield{author}{\bibinfo{person}{Neus Sabater}, \bibinfo{person}{Guillaume
  Boisson}, \bibinfo{person}{Benoit Vandame}, \bibinfo{person}{Paul Kerbiriou},
  \bibinfo{person}{Frederic Babon}, \bibinfo{person}{Matthieu Hog},
  \bibinfo{person}{Remy Gendrot}, \bibinfo{person}{Tristan Langlois},
  \bibinfo{person}{Olivier Bureller}, \bibinfo{person}{Arno Schubert}, {and}
  \bibinfo{person}{Valerie Allie}.} \bibinfo{year}{2017}\natexlab{}.
\newblock \showarticletitle{Dataset and Pipeline for Multi-View Light-Field
  Video}. In \bibinfo{booktitle}{\emph{Proceedings of the IEEE Conference on
  Computer Vision and Pattern Recognition (CVPR) Workshop}}.
\newblock


\bibitem[Schonberger and Frahm(2016)]%
        {schonberger2016colmap}
\bibfield{author}{\bibinfo{person}{Johannes~L. Schonberger} {and}
  \bibinfo{person}{Jan-Michael Frahm}.} \bibinfo{year}{2016}\natexlab{}.
\newblock \showarticletitle{Structure-From-Motion Revisited}. In
  \bibinfo{booktitle}{\emph{Proceedings of the IEEE Conference on Computer
  Vision and Pattern Recognition (CVPR)}}.
\newblock


\bibitem[Sederberg and Parry(1986)]%
        {sederberg1986free}
\bibfield{author}{\bibinfo{person}{Thomas~W. Sederberg} {and}
  \bibinfo{person}{Scott~R. Parry}.} \bibinfo{year}{1986}\natexlab{}.
\newblock \showarticletitle{Free-form Deformation of Solid Geometric Models}.
  In \bibinfo{booktitle}{\emph{Proceedings of the ACM Conference on Computer
  Graphics and Interactive Techniques (SIGGRAPH)}}.
\newblock
\urldef\tempurl%
\url{https://doi.org/10.1145/15922.15903}
\showDOI{\tempurl}


\bibitem[Shaw et~al\mbox{.}(2023)]%
        {shaw2023swags}
\bibfield{author}{\bibinfo{person}{Richard Shaw}, \bibinfo{person}{Jifei Song},
  \bibinfo{person}{Arthur Moreau}, \bibinfo{person}{Michal Nazarczuk},
  \bibinfo{person}{Sibi Catley-Chandar}, \bibinfo{person}{Helisa Dhamo}, {and}
  \bibinfo{person}{Eduardo Perez-Pellitero}.} \bibinfo{year}{2023}\natexlab{}.
\newblock \showarticletitle{SWAGS: Sampling Windows Adaptively for Dynamic 3D
  Gaussian Splatting}.
\newblock \bibinfo{journal}{\emph{arXiv e-prints}}, Article
  \bibinfo{articleno}{arXiv:2312.13308} (\bibinfo{year}{2023}),
  \bibinfo{numpages}{arXiv:2312.13308}~pages.
\newblock
\showeprint[arxiv]{2312.13308}


\bibitem[Shi et~al\mbox{.}(2024)]%
        {shi2024zerorf}
\bibfield{author}{\bibinfo{person}{Ruoxi Shi}, \bibinfo{person}{Xinyue Wei},
  \bibinfo{person}{Cheng Wang}, {and} \bibinfo{person}{Hao Su}.}
  \bibinfo{year}{2024}\natexlab{}.
\newblock \showarticletitle{{ZeroRF}: Fast Sparse View 360${}^\circ$
  Reconstruction with Zero Pretraining}. In
  \bibinfo{booktitle}{\emph{Proceedings of the IEEE/CVF Conference on Computer
  Vision and Pattern Recognition (CVPR)}}.
\newblock


\bibitem[Somraj et~al\mbox{.}(2023)]%
        {somraj2023simplenerf}
\bibfield{author}{\bibinfo{person}{Nagabhushan Somraj},
  \bibinfo{person}{Adithyan Karanayil}, {and} \bibinfo{person}{Rajiv
  Soundararajan}.} \bibinfo{year}{2023}\natexlab{}.
\newblock \showarticletitle{{SimpleNeRF}: Regularizing Sparse Input Neural
  Radiance Fields with Simpler Solutions}. In
  \bibinfo{booktitle}{\emph{Proceedings of the ACM Special Interest Group on
  Computer Graphics and Interactive Techniques - Asia (SIGGRAPH-Asia)}}.
\newblock


\bibitem[Somraj et~al\mbox{.}(2022)]%
        {somraj2022decompnet}
\bibfield{author}{\bibinfo{person}{Nagabhushan Somraj},
  \bibinfo{person}{Pranali Sancheti}, {and} \bibinfo{person}{Rajiv
  Soundararajan}.} \bibinfo{year}{2022}\natexlab{}.
\newblock \showarticletitle{Temporal View Synthesis of Dynamic Scenes through
  3D Object Motion Estimation with Multi-Plane Images}. In
  \bibinfo{booktitle}{\emph{Proceedings of the IEEE International Symposium on
  Mixed and Augmented Reality (ISMAR)}}.
\newblock
\urldef\tempurl%
\url{https://doi.org/10.1109/ISMAR55827.2022.00100}
\showDOI{\tempurl}


\bibitem[Somraj and Soundararajan(2023)]%
        {somraj2023vipnerf}
\bibfield{author}{\bibinfo{person}{Nagabhushan Somraj} {and}
  \bibinfo{person}{Rajiv Soundararajan}.} \bibinfo{year}{2023}\natexlab{}.
\newblock \showarticletitle{{ViP-NeRF}: Visibility Prior for Sparse Input
  Neural Radiance Fields}. In \bibinfo{booktitle}{\emph{Proceedings of the ACM
  Special Interest Group on Computer Graphics and Interactive Techniques
  (SIGGRAPH)}}.
\newblock
\urldef\tempurl%
\url{https://doi.org/10.1145/3588432.3591539}
\showDOI{\tempurl}


\bibitem[Stoll et~al\mbox{.}(2011)]%
        {stoll2011fast}
\bibfield{author}{\bibinfo{person}{Carsten Stoll}, \bibinfo{person}{Nils
  Hasler}, \bibinfo{person}{Juergen Gall}, \bibinfo{person}{Hans-Peter Seidel},
  {and} \bibinfo{person}{Christian Theobalt}.} \bibinfo{year}{2011}\natexlab{}.
\newblock \showarticletitle{Fast Articulated Motion Tracking using a Sums of
  {Gaussians} Body Model}. In \bibinfo{booktitle}{\emph{Proceedings of the IEEE
  International Conference on Computer Vision (ICCV)}}.
\newblock
\urldef\tempurl%
\url{https://doi.org/10.1109/ICCV.2011.6126338}
\showDOI{\tempurl}


\bibitem[Sun et~al\mbox{.}(2022)]%
        {sun2022dvgo}
\bibfield{author}{\bibinfo{person}{Cheng Sun}, \bibinfo{person}{Min Sun}, {and}
  \bibinfo{person}{Hwann-Tzong Chen}.} \bibinfo{year}{2022}\natexlab{}.
\newblock \showarticletitle{Direct Voxel Grid Optimization: Super-Fast
  Convergence for Radiance Fields Reconstruction}. In
  \bibinfo{booktitle}{\emph{Proceedings of the IEEE/CVF Conference on Computer
  Vision and Pattern Recognition (CVPR)}}.
\newblock


\bibitem[Teed and Deng(2020)]%
        {teed2020raft}
\bibfield{author}{\bibinfo{person}{Zachary Teed} {and} \bibinfo{person}{Jia
  Deng}.} \bibinfo{year}{2020}\natexlab{}.
\newblock \showarticletitle{{RAFT}: Recurrent All-Pairs Field Transforms for
  Optical Flow}. In \bibinfo{booktitle}{\emph{Proceedings of the European
  Conference on Computer Vision (ECCV)}}.
\newblock


\bibitem[Uy et~al\mbox{.}(2023)]%
        {uy2023scade}
\bibfield{author}{\bibinfo{person}{Mikaela~Angelina Uy},
  \bibinfo{person}{Ricardo Martin-Brualla}, \bibinfo{person}{Leonidas Guibas},
  {and} \bibinfo{person}{Ke Li}.} \bibinfo{year}{2023}\natexlab{}.
\newblock \showarticletitle{SCADE: NeRFs from Space Carving with
  Ambiguity-Aware Depth Estimates}.
\newblock  (\bibinfo{date}{June} \bibinfo{year}{2023}).
\newblock


\bibitem[Vedula et~al\mbox{.}(2000)]%
        {vedula2000shape}
\bibfield{author}{\bibinfo{person}{S. Vedula}, \bibinfo{person}{S. Baker},
  \bibinfo{person}{S. Seitz}, {and} \bibinfo{person}{T. Kanade}.}
  \bibinfo{year}{2000}\natexlab{}.
\newblock \showarticletitle{Shape and Motion Carving in {6D}}. In
  \bibinfo{booktitle}{\emph{Proceedings of the IEEE Conference on Computer
  Vision and Pattern Recognition (CVPR)}}.
\newblock
\urldef\tempurl%
\url{https://doi.org/10.1109/CVPR.2000.854926}
\showDOI{\tempurl}


\bibitem[Vlasic et~al\mbox{.}(2008)]%
        {vlasic2008articulated}
\bibfield{author}{\bibinfo{person}{Daniel Vlasic}, \bibinfo{person}{Ilya
  Baran}, \bibinfo{person}{Wojciech Matusik}, {and} \bibinfo{person}{Jovan
  Popovi\'{c}}.} \bibinfo{year}{2008}\natexlab{}.
\newblock \showarticletitle{Articulated Mesh Animation from Multi-View
  Silhouettes}. In \bibinfo{booktitle}{\emph{Proceedings of the ACM Conference
  on Computer Graphics and Interactive Techniques (SIGGRAPH)}}.
\newblock
\urldef\tempurl%
\url{https://doi.org/10.1145/1399504.1360696}
\showDOI{\tempurl}


\bibitem[Vlasic et~al\mbox{.}(2009)]%
        {vlasic2009dynamic}
\bibfield{author}{\bibinfo{person}{Daniel Vlasic}, \bibinfo{person}{Pieter
  Peers}, \bibinfo{person}{Ilya Baran}, \bibinfo{person}{Paul Debevec},
  \bibinfo{person}{Jovan Popovi\'{c}}, \bibinfo{person}{Szymon Rusinkiewicz},
  {and} \bibinfo{person}{Wojciech Matusik}.} \bibinfo{year}{2009}\natexlab{}.
\newblock \showarticletitle{Dynamic Shape Capture using Multi-View Photometric
  Stereo}. In \bibinfo{booktitle}{\emph{Proceedings of the ACM Conference on
  Computer Graphics and Interactive Techniques (SIGGRAPH)}}.
\newblock
\urldef\tempurl%
\url{https://doi.org/10.1145/1661412.1618520}
\showDOI{\tempurl}


\bibitem[Wang et~al\mbox{.}(2021)]%
        {wang2021neural}
\bibfield{author}{\bibinfo{person}{Chaoyang Wang}, \bibinfo{person}{Ben
  Eckart}, \bibinfo{person}{Simon Lucey}, {and} \bibinfo{person}{Orazio
  Gallo}.} \bibinfo{year}{2021}\natexlab{}.
\newblock \showarticletitle{Neural Trajectory Fields for Dynamic Novel View
  Synthesis}.
\newblock \bibinfo{journal}{\emph{arXiv e-prints}}, Article
  \bibinfo{articleno}{arXiv:2105.05994} (\bibinfo{year}{2021}),
  \bibinfo{numpages}{arXiv:2105.05994}~pages.
\newblock
\showeprint[arxiv]{2105.05994}


\bibitem[Wang et~al\mbox{.}(2023b)]%
        {wang2023flow}
\bibfield{author}{\bibinfo{person}{Chaoyang Wang},
  \bibinfo{person}{Lachlan~Ewen MacDonald}, \bibinfo{person}{L\'aszl\'o~A.
  Jeni}, {and} \bibinfo{person}{Simon Lucey}.}
  \bibinfo{year}{2023}\natexlab{b}.
\newblock \showarticletitle{Flow Supervision for Deformable NeRF}. In
  \bibinfo{booktitle}{\emph{Proceedings of the IEEE/CVF Conference on Computer
  Vision and Pattern Recognition (CVPR)}}.
\newblock


\bibitem[Wang et~al\mbox{.}(2009)]%
        {wang2009physically}
\bibfield{author}{\bibinfo{person}{Huamin Wang}, \bibinfo{person}{Miao Liao},
  \bibinfo{person}{Qing Zhang}, \bibinfo{person}{Ruigang Yang}, {and}
  \bibinfo{person}{Greg Turk}.} \bibinfo{year}{2009}\natexlab{}.
\newblock \showarticletitle{Physically Guided Liquid Surface Modeling from
  Videos}.
\newblock \bibinfo{journal}{\emph{ACM Transactions on Graphics (TOG)}}
  \bibinfo{volume}{28}, \bibinfo{number}{3} (\bibinfo{date}{July}
  \bibinfo{year}{2009}).
\newblock
\urldef\tempurl%
\url{https://doi.org/10.1145/1531326.1531396}
\showDOI{\tempurl}


\bibitem[Wang et~al\mbox{.}(2023a)]%
        {wang2023tracking}
\bibfield{author}{\bibinfo{person}{Qianqian Wang}, \bibinfo{person}{Yen-Yu
  Chang}, \bibinfo{person}{Ruojin Cai}, \bibinfo{person}{Zhengqi Li},
  \bibinfo{person}{Bharath Hariharan}, \bibinfo{person}{Aleksander Holynski},
  {and} \bibinfo{person}{Noah Snavely}.} \bibinfo{year}{2023}\natexlab{a}.
\newblock \showarticletitle{Tracking Everything Everywhere All at Once}. In
  \bibinfo{booktitle}{\emph{Proceedings of the IEEE/CVF International
  Conference on Computer Vision (ICCV)}}.
\newblock


\bibitem[Wang et~al\mbox{.}(2003)]%
        {wang2003multiscale}
\bibfield{author}{\bibinfo{person}{Zhou Wang}, \bibinfo{person}{Eero~P
  Simoncelli}, {and} \bibinfo{person}{Alan~C Bovik}.}
  \bibinfo{year}{2003}\natexlab{}.
\newblock \showarticletitle{Multiscale structural similarity for image quality
  assessment}. In \bibinfo{booktitle}{\emph{Proceedings of the Asilomar
  Conference on Signals, Systems Computers}}.
\newblock


\bibitem[Wu et~al\mbox{.}(2023)]%
        {wu20234dgs}
\bibfield{author}{\bibinfo{person}{Guanjun Wu}, \bibinfo{person}{Taoran Yi},
  \bibinfo{person}{Jiemin Fang}, \bibinfo{person}{Lingxi Xie},
  \bibinfo{person}{Xiaopeng Zhang}, \bibinfo{person}{Wei Wei},
  \bibinfo{person}{Wenyu Liu}, \bibinfo{person}{Qi Tian}, {and}
  \bibinfo{person}{Wang Xinggang}.} \bibinfo{year}{2023}\natexlab{}.
\newblock \showarticletitle{{4D} Gaussian Splatting for Real-Time Dynamic Scene
  Rendering}.
\newblock \bibinfo{journal}{\emph{arXiv e-prints}}, Article
  \bibinfo{articleno}{arXiv:2310.08528} (\bibinfo{year}{2023}),
  \bibinfo{numpages}{arXiv:2310.08528}~pages.
\newblock
\showeprint[arxiv]{2310.08528}


\bibitem[Wu et~al\mbox{.}(2024)]%
        {wu2024reconfusion}
\bibfield{author}{\bibinfo{person}{Rundi Wu}, \bibinfo{person}{Ben Mildenhall},
  \bibinfo{person}{Philipp Henzler}, \bibinfo{person}{Keunhong Park},
  \bibinfo{person}{Ruiqi Gao}, \bibinfo{person}{Daniel Watson},
  \bibinfo{person}{Pratul~P Srinivasan}, \bibinfo{person}{Dor Verbin},
  \bibinfo{person}{Jonathan~T Barron}, \bibinfo{person}{Ben Poole},
  {et~al\mbox{.}}} \bibinfo{year}{2024}\natexlab{}.
\newblock \showarticletitle{{ReconFusion}: {3D} Reconstruction with Diffusion
  Priors}. In \bibinfo{booktitle}{\emph{Proceedings of the IEEE/CVF Conference
  on Computer Vision and Pattern Recognition (CVPR)}}.
\newblock


\bibitem[Wynn and Turmukhambetov(2023)]%
        {wynn2023diffusionerf}
\bibfield{author}{\bibinfo{person}{Jamie Wynn} {and} \bibinfo{person}{Daniyar
  Turmukhambetov}.} \bibinfo{year}{2023}\natexlab{}.
\newblock \showarticletitle{{DiffusioNeRF}: Regularizing Neural Radiance Fields
  with Denoising Diffusion Models}.
\newblock \bibinfo{journal}{\emph{arXiv e-prints}}, Article
  \bibinfo{articleno}{arXiv:2302.12231} (\bibinfo{year}{2023}),
  \bibinfo{numpages}{arXiv:2302.12231}~pages.
\newblock
\showeprint[arxiv]{2302.12231}


\bibitem[Xing and Chen(2021)]%
        {xing2021temporal}
\bibfield{author}{\bibinfo{person}{Wenpeng Xing} {and} \bibinfo{person}{Jie
  Chen}.} \bibinfo{year}{2021}\natexlab{}.
\newblock \showarticletitle{Temporal-{MPI}: Enabling Multi-Plane Images for
  Dynamic Scene Modelling via Temporal Basis Learning}.
\newblock \bibinfo{journal}{\emph{arXiv e-prints}}, Article
  \bibinfo{articleno}{arXiv:2111.10533} (\bibinfo{year}{2021}),
  \bibinfo{numpages}{arXiv:2111.10533}~pages.
\newblock
\showeprint[arxiv]{2111.10533}


\bibitem[Xiong et~al\mbox{.}(2023)]%
        {xiong2023sparsegs}
\bibfield{author}{\bibinfo{person}{Haolin Xiong}, \bibinfo{person}{Sairisheek
  Muttukuru}, \bibinfo{person}{Rishi Upadhyay}, \bibinfo{person}{Pradyumna
  Chari}, {and} \bibinfo{person}{Achuta Kadambi}.}
  \bibinfo{year}{2023}\natexlab{}.
\newblock \showarticletitle{{SparseGS}: Real-Time 360${}^\circ$ Sparse View
  Synthesis using Gaussian Splatting}.
\newblock \bibinfo{journal}{\emph{arXiv e-prints}}, Article
  \bibinfo{articleno}{arXiv:2312.00206} (\bibinfo{year}{2023}),
  \bibinfo{numpages}{arXiv:2312.00206}~pages.
\newblock
\showeprint[arxiv]{2312.00206}


\bibitem[Yoon et~al\mbox{.}(2020)]%
        {yoon2020novel}
\bibfield{author}{\bibinfo{person}{Jae~Shin Yoon}, \bibinfo{person}{Kihwan
  Kim}, \bibinfo{person}{Orazio Gallo}, \bibinfo{person}{Hyun~Soo Park}, {and}
  \bibinfo{person}{Jan Kautz}.} \bibinfo{year}{2020}\natexlab{}.
\newblock \showarticletitle{Novel View Synthesis of Dynamic Scenes With
  Globally Coherent Depths From a Monocular Camera}. In
  \bibinfo{booktitle}{\emph{Proceedings of the IEEE Conference on Computer
  Vision and Pattern Recognition (CVPR)}}.
\newblock


\bibitem[Yu et~al\mbox{.}(2023)]%
        {yu2023cogs}
\bibfield{author}{\bibinfo{person}{Heng Yu}, \bibinfo{person}{Joel Julin},
  \bibinfo{person}{Zolt{\'a}n~{\'A} Milacski}, \bibinfo{person}{Koichiro
  Niinuma}, {and} \bibinfo{person}{L{\'a}szl{\'o}~A Jeni}.}
  \bibinfo{year}{2023}\natexlab{}.
\newblock \showarticletitle{CoGS: Controllable Gaussian Splatting}.
\newblock \bibinfo{journal}{\emph{arXiv e-prints}}, Article
  \bibinfo{articleno}{arXiv:2312.05664} (\bibinfo{year}{2023}),
  \bibinfo{numpages}{arXiv:2312.05664}~pages.
\newblock
\showeprint[arxiv]{2312.05664}


\bibitem[Zhang et~al\mbox{.}(2018)]%
        {zhang2018unreasonable}
\bibfield{author}{\bibinfo{person}{Richard Zhang}, \bibinfo{person}{Phillip
  Isola}, \bibinfo{person}{Alexei~A Efros}, \bibinfo{person}{Eli Shechtman},
  {and} \bibinfo{person}{Oliver Wang}.} \bibinfo{year}{2018}\natexlab{}.
\newblock \showarticletitle{The Unreasonable Effectiveness of Deep Features as
  a Perceptual Metric}. In \bibinfo{booktitle}{\emph{Proceedings of the IEEE
  Conference on Computer Vision and Pattern Recognition (CVPR)}}.
\newblock


\bibitem[Zhou et~al\mbox{.}(2018)]%
        {zhou2018stereomag}
\bibfield{author}{\bibinfo{person}{Tinghui Zhou}, \bibinfo{person}{Richard
  Tucker}, \bibinfo{person}{John Flynn}, \bibinfo{person}{Graham Fyffe}, {and}
  \bibinfo{person}{Noah Snavely}.} \bibinfo{year}{2018}\natexlab{}.
\newblock \showarticletitle{Stereo Magnification: Learning View Synthesis Using
  Multiplane Images}.
\newblock \bibinfo{journal}{\emph{ACM Transactions on Graphics (TOG)}}
  \bibinfo{volume}{37}, \bibinfo{number}{4} (\bibinfo{date}{July}
  \bibinfo{year}{2018}).
\newblock


\bibitem[Zhu et~al\mbox{.}(2023)]%
        {zhu2023fsgs}
\bibfield{author}{\bibinfo{person}{Zehao Zhu}, \bibinfo{person}{Zhiwen Fan},
  \bibinfo{person}{Yifan Jiang}, {and} \bibinfo{person}{Zhangyang Wang}.}
  \bibinfo{year}{2023}\natexlab{}.
\newblock \showarticletitle{{FSGS}: Real-Time Few-shot View Synthesis using
  Gaussian Splatting}.
\newblock \bibinfo{journal}{\emph{arXiv e-prints}}, Article
  \bibinfo{articleno}{arXiv:2312.00451} (\bibinfo{year}{2023}),
  \bibinfo{numpages}{arXiv:2312.00451}~pages.
\newblock
\showeprint[arxiv]{2312.00451}


\bibitem[Zitnick et~al\mbox{.}(2004)]%
        {zitnick2004high}
\bibfield{author}{\bibinfo{person}{C.~Lawrence Zitnick},
  \bibinfo{person}{Sing~Bing Kang}, \bibinfo{person}{Matthew Uyttendaele},
  \bibinfo{person}{Simon Winder}, {and} \bibinfo{person}{Richard Szeliski}.}
  \bibinfo{year}{2004}\natexlab{}.
\newblock \showarticletitle{High-Quality Video View Interpolation using a
  Layered Representation}.
\newblock \bibinfo{journal}{\emph{ACM Transactions on Graphics (TOG)}}
  \bibinfo{volume}{23}, \bibinfo{number}{3} (\bibinfo{date}{August}
  \bibinfo{year}{2004}), \bibinfo{pages}{600--608}.
\newblock
\urldef\tempurl%
\url{https://doi.org/10.1145/1015706.1015766}
\showDOI{\tempurl}


\end{thebibliography}
